\documentclass[nohyperref]{article}

\usepackage{microtype}
\usepackage{graphicx}
\usepackage{subfigure}
\usepackage{booktabs} 
\usepackage{multirow}

\usepackage{hyperref}

\usepackage[accepted]{icml2023}

\usepackage{amsmath}
\usepackage{amssymb}
\usepackage{mathtools}
\usepackage{amsthm}

\usepackage[capitalize,noabbrev]{cleveref}

\DeclareMathOperator*{\argmin}{arg\,min}

\theoremstyle{plain}

\theoremstyle{definition}

\theoremstyle{remark}

\usepackage[textsize=tiny]{todonotes}

\icmltitlerunning{Learning Perturbations to Explain Time Series Predictions}

\begin{document}

\twocolumn[
\icmltitle{Learning Perturbations to Explain Time Series Predictions}



\icmlsetsymbol{equal}{*}

\begin{icmlauthorlist}
\icmlauthor{Joseph Enguehard}{babylon,skippr}
\end{icmlauthorlist}

\icmlaffiliation{babylon}{Babylon Health, 1 Knightsbridge Grn, London SW1X 7QA United Kingdom}

\icmlaffiliation{skippr}{Skippr, 99 Milton Keynes Business Centre, Milton Keynes MK14 6GD United Kingdom}

\icmlcorrespondingauthor{Joseph Enguehard}{joseph@skippr.com}

\icmlkeywords{Machine Learning, ICML}

\vskip 0.3in
]



\printAffiliationsAndNotice{}  

\begin{abstract}
Explaining predictions based on multivariate time series data carries the additional difficulty of handling not only multiple features, but also time dependencies. It matters not only what happened, but also when, and the same feature could have a very different impact on a prediction depending on this time information. Previous work has used perturbation-based saliency methods to tackle this issue, perturbing an input using a trainable mask to discover which features at which times are driving the predictions. However these methods introduce fixed perturbations, inspired from similar methods on static data, while there seems to be little motivation to do so on temporal data. In this work, we aim to explain predictions by learning not only masks, but also associated perturbations. We empirically show that learning these perturbations significantly improves the quality of these explanations on time series data.
\end{abstract}

\section{Introduction}
\label{introduction}

\begin{figure}[ht]
\begin{center}
\centerline{\includegraphics[width=0.8\columnwidth]{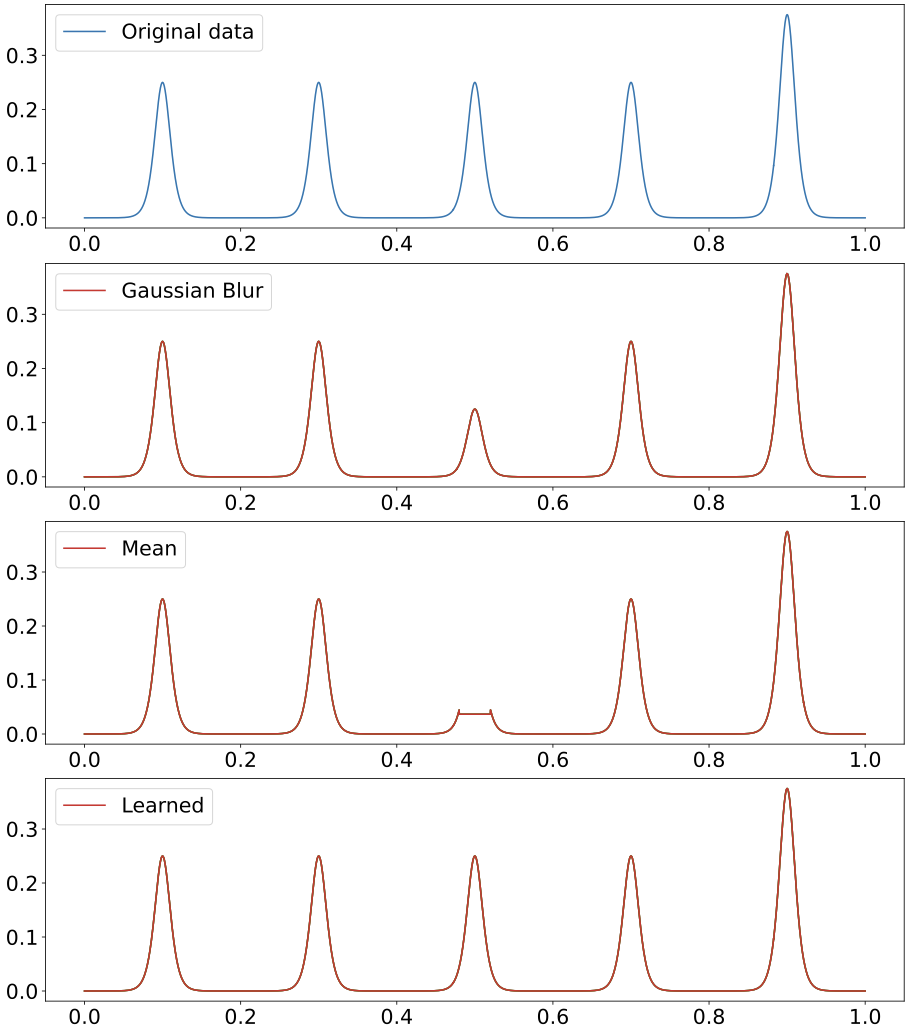}}
\caption{\textbf{Illustration of different perturbations} on a time-series (first plot). We aim to evaluate the importance of the third spike here. It is very likely that this spike is unimportant, as it is a regularity in the data, and only the last, larger spike, could matter. However, using a Gaussian blur (second plot) or replacing the original data with an average (third plot) changes the input significantly, which could lead an explanation method to wrongly state that this spike is important. Our learned perturbation (last plot) should on the other hand replace the explained data with another spike, leading to little difference in the output, and therefore correctly stating that this data is unimportant.}
\label{fig:illustration}
\end{center}
\end{figure}

Explaining neural networks predictions has received increasing attention, as these models become more embedded in many decision processes. It is indeed important to understand why such models, which are intrinsically difficult to explain and are often qualified as ``black boxes", made a specific prediction. This information is crucial to assess the fairness of an algorithm in impactful situations, such as when providing a medical diagnosis \citep{ahuja2019impact} or computing a credit score \citep{moscato2021benchmark}. Explaining a deep neural network's predictions is also important to build trust for the users of such technologies. In the medical field, this has been proven to be a crucial step in building this trust \citep{larosa2018impacts}.

As a result, multiple methods to explain why a model made a specific prediction have been recently developed. Some of these methods, such as Lime \citep{ribeiro2016should} or Shap \citep{lundberg2017unified} explain a model's predictions by approximating it locally using a transparent method, in this case a weighted linear regression. Other methods aim to specifically explain a neural network's output by leveraging the back-propagation algorithm to compute the gradient of an output w.r.t. an input \citep{simonyan2013deep}. A feature with an associated high gradient can indeed be interpreted as important, the sign of this gradient indicating if this feature influences the prediction positively or negatively. Several variations of this method have also been developed \citep{sundararajan2017axiomatic, shrikumar2017learning}.

Another important class of explanation methods is called \textit{perturbation-based}. These methods consist in perturbing a feature or a group of features, and measuring how the resulting prediction changes. A greater change indicates a higher importance of the perturbed features. Such methods include Occlusion \citep{zeiler2014visualizing}, which masks features to estimate their importance. Extremal Masks \citep{fong2017interpretable, fong2019understanding} is another perturbation-based method, which learns a mask used to perturb the input. We present this method in more detail in the next section.

However, while many explanation methods have been proposed to explain a neural network, few have been developed to handle multivariate time series data. Yet, this type of data is especially important in the medical field, where the data can be a list of timestamped medical events, or of vitals measurements. There is therefore a need to adapt explanation methods to handle this temporal element. These adaptations currently include RETAIN \citep{choi2016retain}, an attention-based model which learns this attention over features and time, or FIT \cite{tonekaboni2020went}, which estimates the importance of features over time by quantifying the shift in the predictive distribution. Another method, DynaMask \citep{crabbe2021explaining}, adapts perturbation-based methods to multivariate time-series. We will present and discuss this method further in the next section.

In this work, we aim to further adapt perturbation-based methods to multivariate time-series driven with the following insight. In the works of \citet{fong2017interpretable} and \citet{crabbe2021explaining}, while the mask is learned, the perturbation induced by this mask is fixed. For instance, \citet{fong2017interpretable} replaces a feature with a Gaussian blur (a weighted average of data around the feature) depending on the value of the feature's mask: the lower this value, the higher the amount of blur. \citet{crabbe2021explaining} adapts this method by blurring the data temporally. This method seems reasonable with images, where information can be assumed to be local, which explains why convolutional neural networks (CNNs), which have a limited filter size, still perform very well on such data. However, multivariate time-series can have long-term dependencies which makes it less obvious to use a temporal Gaussian blur as the perturbation. Instead of replacing a masked feature with a local average, we might want to replace it using data further away in time. But then, how should we choose the correct perturbation formula?

This calls to replace fixed perturbations with learnable ones. In this work, we present such a method\footnote{An implement of this work can be found at \url{https://github.com/josephenguehard/time_interpret}} and empirically show that it significantly improves the quality of the explanations, evaluated on both synthetic and real-world data. This study is organised as follows. We first present in more detail the methods of \citet{fong2017interpretable} and \citet{crabbe2021explaining} in the next section. We then present our method in the following one. We conduct several experiments in the next section, designed to compare our method with several baselines, and we provide elements of discussion in the last section.
\section{Background Work}
\label{related}

In this section, we describe in more detail two methods: one developed by \citet{fong2017interpretable} and its adaptation to time series by \citet{crabbe2021explaining}.

\citet{fong2017interpretable} propose a perturbation-based method which is defined as following. A trainable mask, with values restricted between 0 and 1, is used to generate perturbed data, which is then passed to the neural network to be explained in order to compute predictions. This mask can then be trained in two different manners, that the authors call the \textit{deletion game} and the \textit{preservation game}. In the deletion game, we aim to mask as little data as possible, while trying to reduce as much as possible the predictions, on the targeted class, of the perturbed data, compared with the original predictions. This objective can be defined as, for a model $f: \mathbb{R}^n \to \mathbb{R}^p$, a mask $\textbf{m} \in [0, 1]^n$, an input $\textbf{x} \in \mathbb{R}^n$ and a perturbation $\Phi(\textbf{x}, \textbf{m}): \mathbb{R}^n \times [0, 1]^n \to \mathbb{R}^n$:

\begin{equation}
    \argmin_{\textbf{m} \in [0, 1]^n} \lambda ||\textbf{1} - \textbf{m}||_1 - \mathcal{L}(f(\textbf{x}), f(\Phi(\textbf{x}, \textbf{m})))
\label{eq:deletion}
\end{equation}

The value n represents the input dimension, and $\lambda$ is a hyperparameter balancing both goals. 

Secondly, in the preservation game, we aim to retain the least amount of data that will preserve the closest predictions compared with the original ones on the targeted class. This objective can be defined as:

\begin{equation}
     \argmin_{\textbf{m} \in [0, 1]^n} \lambda ||\textbf{m}||_1 + \mathcal{L}(f(\textbf{x}), f(\Phi(\textbf{x}, \textbf{m})))
\label{eq:preservation}
\end{equation}

Moreover, the perturbation $\Phi(\textbf{x}, \textbf{m})$ is fixed given an input and a mask. \citet{fong2017interpretable} define several strategies \footnote{Unintuitively, the original data is masked when \textbf{m} = 0. We kept this notation as it is used in both \citet{fong2017interpretable} and \citet{crabbe2021explaining}.}:

\begin{equation}
    \Phi(\textbf{x}, \textbf{m}) = 
    \begin{cases}
        \textbf{m} \times \textbf{x} + (\textbf{1} - \textbf{m}) \times \mu_0 \\
        \textbf{m} \times \textbf{x} + (\textbf{1} - \textbf{m}) \times \nu \\
        \int g_{\sigma_0 \times (1 - \textbf{m})} (\textbf{y} - \textbf{x}) \, d\textbf{y}
    \end{cases}
\end{equation}

The first strategy corresponds to replacing the original masked value $\textbf{x}$ with an average $\mu_0$, the second one consisting in replacing this value with Gaussian noise: $ \nu \sim \mathcal{N}(0, 1)$ and the last replaces it with a Gaussian blur $g_{\sigma}$ around $\textbf{x}$, given a maximum std $\sigma_0$. \citet{fong2017interpretable} also add some regularisation to ensure the perturbation to be more natural in the context of computer vision, but we leave it out, as it is further away from our topic.

While this method was developed to explain predictions based on images, \citet{crabbe2021explaining} adapted it to multivariate time series. They propose as a result a method they call DynaMask, as the learned mask contains in this case a time dimension. The input space is now $\mathbb{R}^{\mathrm{T} \times n}$, and we consider similarly a neural network $f$ and a target class $c$ such as: $f_c(\textbf{x}): \mathbb{R}^{\mathrm{T} \times n} \to \mathbb{R}$. Therefore, the mask $\textbf{m} \in \mathbb{R}^{\mathrm{T} \times n}$ and the input $\textbf{x} \in \mathbb{R}^{\mathrm{T} \times n}$ are also defined on this input space.

The main contribution of \citet{crabbe2021explaining} is to then adapt the perturbation operator $\Phi$ to account for this temporal information. They also introduce three strategies:

\begin{equation}
\Phi(\textbf{x}, \textbf{m})_{t, i} = 
\begin{cases}
    m_{t, i} \times x_{t, i} + (1 - m_{t, i}) \times \mu_{t, i} \\
    m_{t, i} \times x_{t, i} + (1 - m_{t, i}) \times \mu^p_{t, i} \\
    \frac{\sum_{t'=1}^\textrm{T} x_{t', i} \times g_{\sigma(m_{t, i})}(t - t')}{\sum_{t'=1}^\textrm{T} g_{\sigma(m_{t, i})}(t - t')}
\end{cases}
\label{eq:perturbations}
\end{equation}

Where $\mu_{t, i}$ is an average of $\textbf{x}_{:,i}$ over a window W around $t$:

\begin{equation}
    \mu_{t, i} = \frac{1}{2 \textrm{W} + 1} \sum_{t'=t-\textrm{W}}^{t+\textrm{W}} x_{t', i}
\end{equation}

and $\mu^p_{t, i}$ is an average of $\textbf{x}_{:,i}$ over a past element up to $t$:

\begin{equation}
    \mu^p_{t, i} = \frac{1}{\textrm{W} + 1} \sum_{t'=t-\textrm{W}}^t x_{t', i}
\end{equation}

Finally, the last perturbation is a temporal Gaussian blur:

\begin{equation}
    g_{\sigma(m_{t, i})}(t) = \exp(-\frac{t^2}{2\sigma^2}) ; \, \sigma(\textbf{m}) = \sigma_{\textrm{max}} (\textbf{1} - \textbf{m})
\end{equation}

\citet{crabbe2021explaining} uses these perturbations in a preservation game, which aims to mask the maximum amount of data while keeping close predictions compared with the originals. They also leverage further work of \citet{fong2017interpretable}: \citet{fong2019understanding}, which replaces Equations \ref{eq:deletion} and \ref{eq:preservation} with an area constraint. In the preservation mode (the deletion mode can be adapted similarly), the regulation $\lambda ||\textbf{m}||_1$ in Equation \ref{eq:preservation} is replaced with: $\lambda_a(\textbf{m}) = ||\textrm{vecsort}(\textbf{m}) - \textbf{r}_a||^2$, where $a$ is a number between 0 and 1, $\textrm{vecsort}(\textbf{m})$ sorts the values of $\textbf{m}$ from lowest to largest, and $\textbf{r}_a$ is a vector containing $(1 - a) \times T \times n$ zeros followed by $a \times T \times n$. As a result, this constraint allows the user to define how much of the data should be masked. In practice, \citet{crabbe2021explaining} use $a$ as a hyperparameter, which is tuned for each data point to be explained.
\section{Method}
\label{method}

\begin{figure*}[ht]
\begin{center}
\centerline{\includegraphics[width=0.9\textwidth]{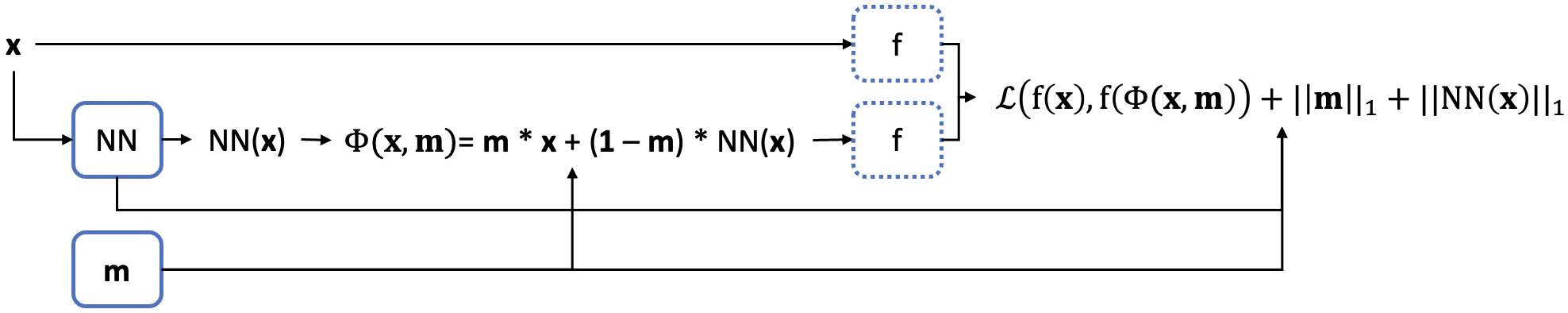}}
\caption{\textbf{Illustration of our method.} The input is passed through a neural network NN to create a perturbation. A mask \textbf{m} is then used to balance the amount of perturbed data: NN(\textbf{x}) and unperturbed data: \textbf{x}, resulting in $\Phi(\textbf{x}, \textbf{m})$. Both \textbf{x} and $\Phi(\textbf{x}, \textbf{m})$ are then passed through the model to explain f. Learnable parameters (\textbf{m} and NN(\textbf{x})) are presented in continuous boxes, while fixed parameters (the model f) are presented in dashed boxes. The objective of this method is to keep the predictions of the perturbed data as close as possible to the original ones, while masking as much data as possible and to keep the perturbations NN(\textbf{x}) as sparse as possible. The overall goal is therefore to identify which features are salient enough to be sufficient to recover the original predictions, when all other features are masked.}
\label{fig:method}
\end{center}
\end{figure*}

While \citet{crabbe2021explaining} propose temporal perturbations as adaptations of the ones defined by \citet{fong2017interpretable} in a computer vision context, these perturbations are kept fixed and local. They are indeed defined either as a moving average perturbation, or as a temporal Gaussian blur. However, temporal data is often characterised by long-term dependencies, and local information can therefore be insufficient to determine the importance of a feature at a particular time. For instance, temporal data can include repetitive patterns, as illustrated on Figure \ref{fig:illustration}, which cannot be taken account using only temporally local information. Moreover, while the perturbations proposed by \citet{crabbe2021explaining} do include the possibility to include data further away in time, by tuning the size of the window W, or the parameter $\sigma_{\max}$ for the Gaussian blur, it is not clear how to choose such parameters nor how this would solve the issue of long term patterns.

This insight calls for a generalised perturbation, which can be tuned to the data we are aiming to explain. A first idea would be to directly learn this perturbation $\Phi(\textbf{x})$, without needing a mask, by optimizing a function similar to Equation \ref{eq:preservation}. However, this method is problematic as it gives too much liberty to the perturbation model. Indeed, such a model, incentivised to output sparse explanations, could compress the data information into a small part of the input space, stating that this part is important while the rest is uninformative. On the contrary, we need to constrain the perturbation operator to explain each part of the input data without changing or moving it.

To overcome this difficulty, we take inspiration from the perturbation operators of \citet{crabbe2021explaining} in Equation \ref{eq:perturbations}. These perturbations are generally defined as $\textbf{m} \times \textbf{x} + (1 - \textbf{m}) \times \mu(\textbf{x})$, where $\mu(\textbf{x})$ is a function of the input. In this work, we propose to replace these fixed functions with a neural network (NN), and to train it in combination with the mask. Our perturbation is therefore defined as:

\begin{equation}
\begin{split}
    \Phi(\textbf{x}, \textbf{m}) &= \textbf{m} \times \textbf{x} + (\textbf{1} - \textbf{m}) \times \textrm{NN}(\textbf{x})\\
    \textbf{0} &\leq \textbf{m}\leq \textbf{1}
\end{split}
\label{eq:method}
\end{equation}

By keeping $\textbf{m}$ between 0 and 1, we constrain the mask to only learn how important each feature is. Moreover, we can see that this equation can be interpreted as a generalisation of the perturbations from \citet{crabbe2021explaining} defined in Equation \ref{eq:perturbations}. The neural network in the second component of Equation \ref{eq:method} can indeed, after training, output a Gaussian blur or an average of \textbf{x} over a window.

In practice, we want to model $\textrm{NN}(\textbf{x})$ as a weighted sum of $x_{t, i}, \, t \in \{1, ..., \textrm{T} \}$. As a result, we choose this model to be a bidirectional GRU \citep{cho2014properties}. This would correspond to a general form of a Gaussian blur or a window around each element $x_{t,i}$. We also compare this choice, in the experiment section, with a unidirectional GRU, which would be closer to the $\mu^p_{t,i}$ average in \citet{crabbe2021explaining}.

As in \citet{crabbe2021explaining}, we define the objective of the mask and the GRU combined as a preservation game, aiming to mask as much data as possible while keeping the closest predictions as possible to the original ones. Our objective is therefore:

\begin{equation}
    \argmin_{\textbf{m}, \Theta \in \textrm{NN}} \lambda ||\textbf{m}||_1 + \mathcal{L}(f(\textbf{x}), f(\Phi(\textbf{x}, \textbf{m})))
\label{eq:optim}
\end{equation}

where $\Theta$ represents the parameters of the neural network, and $\mathcal{L}$ represents a loss between the original and the perturbed predictions. This loss can be for instance a mean square error for regression tasks, or a cross-entropy loss for classification tasks.

One issue that can arise from this objective is that the neural network can be rewarded to mimic the original $\textbf{x}$ data. Indeed, we can see from Equation \ref{eq:method} that, if $\textbf{m} = \textbf{0}$, then $\Phi(\textbf{x}, \textbf{m}) = \textrm{NN}(\textbf{x})$. Moreover, if $\textrm{NN}(\textbf{x}) \approx \textbf{x}$, the objective defined in Equation \ref{eq:optim} is approximately zero. To prevent this behavior, we modify Equation \ref{eq:optim} with the following one:

\begin{equation}
    \argmin_{\textbf{m}, \Theta \in \textrm{NN}} \lambda_1 ||\textbf{m}||_1 + \lambda_2 ||\textrm{NN}(\textbf{x})||_1 + \mathcal{L}(f(\textbf{x}), f(\Phi(\textbf{x}, \textbf{m})))
\label{eq:optim_modif}
\end{equation}

In Equation \ref{eq:optim_modif}, we therefore force the perturbations to be minimal, being not null only when there is an incentive to do so. Indeed, in Equation \ref{eq:preservation}, there is a balance on $\Phi$: $||\textbf{m}||_1$ tends to make $\Phi$ uninformative, while $\mathcal{L}$ does the opposite. Equation \ref{eq:optim_modif} differs in that sense from Equation \ref{eq:preservation}, as $||\textbf{m}||_1$  tends to make $\Phi$ close to NN(\textbf{x}), which is not necessarily uninformative. To entice NN(\textbf{x}) to be uninformative, we add the loss $||NN(\textbf{x})||_1$, using zero as a prior. Therefore, breaking down the objective of Equation \ref{eq:optim_modif}, we have:

\begin{itemize}
    \item $||\textbf{m}||_1$ induces $\Phi(\textbf{x})$ to be close to $\textrm{NN}(\textbf{x})$
    \item $||\textrm{NN}(\textbf{x})||_1$ induces $\Phi(\textbf{x})$ to be close to \textbf{0} (uninformative)
    \item $\mathcal{L}$ induces $f(\Phi(\textbf{x}, \textbf{m}))$ to be close to $f(\textbf{x})$ (informative)
\end{itemize}

We also set $\lambda_1 = \lambda_2 = 1$ in our experiments, while an ablation study on the choice of these hyperparameters can be found in Section \ref{experiments} and Appendix \ref{app:lambda}.

Moreover, contrary to \citet{crabbe2021explaining}, we do not use an area constraint $||\textrm{vecsort}(\textbf{m}) - \textbf{r}_a||^2$, as it is not clear how to choose the hyperparameter $a$ on usually complex data. In practice, \citet{crabbe2021explaining} tune this hyperparameter, which is computationally expensive, as it requires to train multiple masks. We propose instead to directly train our model using Equation \ref{eq:optim_modif}.

\newpage
\section{Experiments}
\label{experiments}

Following \citet{tonekaboni2020went} and \citet{crabbe2021explaining}, we perform experiments on two datasets: a synthetic one, generated using a Hidden Markov model, and a real-world one, MIMIC-III \citep{johnson2016mimic}.

\subsection{Hidden Markov model experiment}

We generate data using a 2-state hidden Markov model (HMM), closely following \citet{crabbe2021explaining}. The state $s_t$ can therefore be either 0 or 1, and we generate 200 states: $t \in [1: 200]$. Moreover, the input vector has three features, generated according to the current state: $\textbf{x}_t \sim \mathcal{N}(\boldsymbol{\mu}_{s_t}, \boldsymbol{\Sigma}_{s_t})$. The label $y_t$ is generated only using the last two features, the first one being irrelevant. The choice of which feature is used to generate the label depends on the state:

\begin{equation}
\begin{split}
    y_t &\sim (1 + \exp(x_{2, t})^{-1}) \mspace{10mu} \textrm{if} \mspace{10mu} s_t = 0\\
    y_t &\sim (1 + \exp(x_{3, t})^{-1}) \mspace{10mu} \textrm{if} \mspace{10mu} s_t = 1
\end{split}
\end{equation}

Please refer to \citet{crabbe2021explaining} for more details on this dataset, in particular in the choice of $\boldsymbol{\mu}_{s_t}$ and $\boldsymbol{\Sigma}_{s_t}$. 

We generate 1000 time series using this method, and train a one-layer GRU \citep{cho2014properties} neural network to predict $y_t$ using $\textbf{x}_t$, which we aim to explain.

As we know the true salient features with this dataset, we evaluate our explanation methods by comparing the similarity between salient features produced by each method and the ground truth. To do so, we use standard classification metrics: area under recall (AUR) and area under precision (AUP). We also use two metrics introduced by \citet{crabbe2021explaining}: Information: $I_{\textbf{m}}(\textbf{a}) = - \sum_{(t,i) \in \textbf{a}} \ln(1 - m_{t,i})$ which is analogous to the Shannon information content. A higher value indicates a more informative mask. The second metric is the mask entropy: $S_{\textbf{m}}(\textbf{a}) = - \sum_{(t, y) \in \textbf{a}} m_{t, i} \ln m_{t, i} + (1 - m_{t, i}) \ln(1 - m_{(t, i)})$ which is analogous to Shannon entropy. In both metrics, $\textbf{a}$ corresponds to the true salient features.

We compare our method with the following ones: DeepLift \citep{shrikumar2017learning}, DynaMask \citep{crabbe2021explaining}, Integrated Gradients (IG) \citep{sundararajan2017axiomatic}, GradientShap \citep{lundberg2017unified}, Fit \citep{tjoa2020survey}, Lime \citep{ribeiro2016should}, Augmented Occlusion \citep{tonekaboni2020went}, Occlusion \citep{zeiler2014visualizing} and Retain \citep{choi2016retain}. Furthermore, our method uses a bidirectional GRU for the perturbation model.

\begin{table}[t]
	\centering
	\resizebox{0.48\textwidth}{!}{%
	\begin{tabular}{lcccc}
		\toprule
		\textbf{Method} & AUP $\uparrow$ & AUR $\uparrow$ & I $\uparrow$ & S $\downarrow$ \\
		\midrule
        DeepLift            & 0.920 (0.019) & 0.454 (0.011) & 359 (9.55) & 145 (0.949) \\
        DynaMask            & 0.711 (0.020) & 0.763 (0.026) & 954 (50.0) & 45.4 (0.781) \\
		IG                  & 0.918 (0.019) & 0.454 (0.011) & 359 (11.6) & 146 (0.871) \\
        GradientShap        & 0.849 (0.030) & 0.414 (0.015) & 335 (14.8) & 138 (2.44) \\
        Fit                 & 0.421 (0.013) & 0.549 (0.017) & 436 (22.7) & 164 (2.79) \\
        Lime                & \textbf{0.932} (0.017) & 0.438 (0.008) & 347 (8.46) & 143 (1.47) \\
        Occlusion           & 0.866 (0.032) & 0.393 (0.006) & 322 (14.6) & 137 (1.90) \\
        Aug Occlusion       & 0.755 (0.043) & 0.388 (0.025) & 364 (9.02) & 165 (1.42) \\
        Retain              & 0.645 (0.088) & 0.334 (0.013) & 206 (21.2) & 138 (5.85) \\
		\midrule
		Ours         & 0.885 (0.030) & \textbf{0.781} (0.013) & \textbf{1536} (79.0) & \textbf{34.1} (3,70) \\
		\bottomrule
	\end{tabular}%
	}
	\caption{Results of each explanation method compared with ours. For each metric, $\uparrow$ indicates that higher is better, and $\downarrow$ that lower is better. Mean and std are reported over 5 folds.}
	\label{tab:results_hmm}
\end{table}

We present our results in Table \ref{tab:results_hmm}. These results\footnote{In Tables \ref{tab:results_hmm} and \ref{tab:results_mimic_avg}, some results differ from \citet{crabbe2021explaining} due to a few issues in their original implementation. Please refer to issues 4, 8 and 9 in \url{https://github.com/JonathanCrabbe/Dynamask/issues}.} show that, although our method performs slightly lower than some baselines in terms of AUP, it significantly outperforms all other methods by every other metric. In particular, while it slightly outperforms DynaMask in terms of AUR, it yields better results in terms of AUP, Information and Entropy. These results therefore seem to indicate that using learnable perturbations should be preferred compared with fixed one when explaining predictions based on multivariate time series data.

\paragraph{Ablation study on the lambdas.} We perform here an ablation study to determine which values of $\lambda_1$ and $\lambda_2$ should be used in Equation~\ref{eq:optim_modif}. We therefore run our experiment using various values of $\lambda_1$ and $\lambda_2$. We report our results on Table \ref{tab:ablation_lambda_hmm}.

\begin{table}[ht]
	\centering
    \resizebox{0.48\textwidth}{!}{%
	\begin{tabular}{lc|ccccc}
		\toprule
        \multicolumn{2}{c}{} & \multicolumn{5}{c}{$\lambda_1$} \\
        \multicolumn{2}{c}{} & 0.01 & 0.1 & 1 & 10 & 100 \\
        \cmidrule{3-7}
        \multirow{5}{*}{$\lambda_2$} & 
        0.01 & 0.51 - 0.81 & 0.76 - 0.44 & 0.78 - 0.09 & 0.35 - 0.17 & 0.39 - 0.18 \\
        & 0.1 & 0.51 - 0.91 & 0.65 - 0.83 & 0.95 - 0.08 & 0.32 - 0.16 & 0.37 - 0.20 \\
        & 1 & 0.51 - 0.89 & 0.63 - 0.83 & 0.89 - 0.75 & 0.30 - 0.16 & 0.35 - 0.18 \\
        & 10 & 0.48 - 0.90 & 0.65 - 0.83 & 0.89 - 0.74 & 0.99 - 0.26 & 0.41 - 0.19 \\
        & 100 & 0.49 - 0.90 & 0.65 - 0.84 & 0.90 - 0.74 & 0.99 - 0.27 & 0.37 - 0.17 \\
		\bottomrule
	\end{tabular}%
    }
	\caption{Influence of $\lambda_1$ and $\lambda_2$ from Equation \ref{eq:optim_modif} on the results of the HMM experiment. For each pair of parameters, 2 values are reported: AUP - AUR. The average result over 5 runs is reported.}
	\label{tab:ablation_lambda_hmm}
\end{table}

This table show that first $\lambda_1$ needs to be close to 1 to yield good results. Indeed, a low value means lower regularisation, therefore retaining a lot of unimportant features. A high value, on the other hand, forces m to be mostly 0, yielding most features to be considered unimportant.
Moreover, $\lambda_2$ needs to be at least 1 to force NN(x) to learn uninformative perturbations. Otherwise, there is only a weak mechanism to prevent NN from producing an output similar to x.

\paragraph{Learning perturbations as a deletion game.} We also explore here learning perturbation using Equation \ref{eq:deletion}, masking as little data as possible while changing the model's predictions as much as possible. However, we cannot directly use Equation \ref{eq:deletion} for two reasons. First, the term $- \mathcal{L}(f(\textbf{x}), f(\Phi(\textbf{x}, \textbf{m})))$ is hard to optimize, as it entices $f(\Phi(\textbf{x}, \textbf{m}))$ to be ``far" from $f(\textbf{x})$ while there is no clarity on what ``far" should be here. For this reason, we replace this objective with $\mathcal{L}(f(\textbf{0}), f(\Phi(\textbf{x}, \textbf{m})))$, enticing the predictions to be close to predictions made using \textbf{0}, uninformative, as an input. Second, we need to add the term $||\textrm{NN}(\textbf{x})||_1$ in the loss. This results in the following objective:

\begin{equation}
    \argmin_{\textbf{m}, \Theta \in \textrm{NN}} \lambda_1 ||\textbf{1} - \textbf{m}||_1 + \lambda_2 ||\textrm{NN}(\textbf{x})||_1 + \mathcal{L}(f(\textbf{0}), f(\Phi(\textbf{x}, \textbf{m})))
\label{eq:deletion_modif}
\end{equation}

We present our results on Table~\ref{tab:loss_study_hmm}, comparing the preservation and the deletion modes. While the second setting outperforms the first one in terms of AUR, it performs poorly according to every other metrics. This might be due to the use of $\mathcal{L}(f(\textbf{0}), f(\Phi(\textbf{x}, \textbf{m})))$, which amounts to learning a ``change" in the predictions. This is a less straightforward objective compared with the preservation mode, which aims to retain the original predictions.

\begin{table}[ht]
	\centering
	\resizebox{0.48\textwidth}{!}{%
	\begin{tabular}{lcccc}
		\toprule
		\textbf{Mode} & AUP $\uparrow$ & AUR $\uparrow$ & I $\uparrow$ & S $\downarrow$ \\
		\midrule
        Preservation         & \textbf{0.885} (0.030) & 0.781 (0.013) & \textbf{1536} (79.0) & \textbf{34.1} (3,70) \\
        Deletion            & 0.346 (0.0034) & \textbf{0.863} (0.012) & 1079 (41.5) & 68.0 (5.07) \\
		\bottomrule
	\end{tabular}%
	}
	\caption{Comparison of using the preservation mode vs deletion mode on the HMM experiment. The average result over 5 runs is reported.}
	\label{tab:loss_study_hmm}
\end{table}

\subsection{MIMIC-III experiment}

We evaluate our method on the real-world MIMIC-III dataset, following the works of \citet{tonekaboni2020went} and \citet{crabbe2021explaining}. MIMIC-III consists of patients in intensive-care units (ICU), for which a number of vital signs and lab test results have been regularly measured. The task is here to predict in-hospital mortality of each patient based on 48 hours of data, discretised over each hour. Missing values are imputed using the previous available ones. If there is no previous feature, a standard value is imputed.

We train a one layer GRU with a hidden size of 200 to predict this in-hospital mortality, and we aim to explain this model. In this dataset, the true salient features are unknown, and we need to provide different metrics to evaluate our method. Following \citet{crabbe2021explaining}, we compare the original predictions to ones where a certain proportion of the features have been masked. We replace masked features either with an average over time of this feature: $\overline{x}_{t,i} = \frac{1}{T}\sum_{t} x_{t, i}$, where T = 48 (hours) or with zeros: $\overline{x}_{t,i} = 0$. We use two metrics proposed by \citet{crabbe2021explaining}, and we also draw from the work of \citet{shrikumar2017learning} and \citet{deyoung2019eraser} and propose three additional metrics. These resulting four metrics are then:

\begin{itemize}
    \item \textbf{Accuracy} (Acc): We mask the most salient features and compute the resulting accuracy using this masked data. A lower accuracy means that important features to make accurate predictions have been removed. Therefore, lower is better with this metric.
    \item \textbf{Cross-Entropy} (CE): We mask the most salient features and compute the cross-entropy between predictions made with this masked data with the original one. A higher value indicates that the predictions have more significantly changed and that important features have been removed. Higher is better with this metric.
    \item \textbf{Comprehensiveness} (Comp): We mask the most salient features and compute the average change of the predicted class probability compared with the original one. Higher is better with this metric.
    \item \textbf{Sufficiency} (Suff): We only keep the most salient features, and compute the average change of the predicted class probability compared with the original one. Lower is better with this metric.
\end{itemize}

Similar to our previous experiment, we use a bidirectional GRU as our perturbation model. We compare our method against DeepLift \citep{shrikumar2017learning}, DynaMask \citep{crabbe2021explaining}, Integrated Gradients (IG) \citep{sundararajan2017axiomatic}, GradientShap \citep{lundberg2017unified}, Lime \citep{ribeiro2016should}, Augmented Occlusion \citep{tonekaboni2020went}, Occlusion \citep{zeiler2014visualizing} and Retain \citep{choi2016retain}. 

\begin{table}[t]
	\centering
	\resizebox{0.48\textwidth}{!}{%
	\begin{tabular}{lcccc}
		\toprule
		\textbf{Method} & Acc $\downarrow$ & Comp $\uparrow$ & CE $\uparrow$ & Suff $\downarrow$ \\
		\midrule
        DeepLift            & 0.988 (0.002) & -4.36E-4 (0.001) & 0.097 (0.006) & 2.86E-3 (0.001) \\
        DynaMask            & 0.990 (0.001) & 2.21E-4 (0.001) & 0.097 (0.005) & 2.99E-3 (0.001) \\
		IG                  & 0.988 (0.003) & 2.24E-4 (0.002) & 0.098 (0.006) & 2.21E-3 (0.001) \\
        GradientShap        & 0.987 (0.004) & -2.19E-3 (0.001) & 0.095 (0.006) & 3.99E-3 (0.001) \\
        Lime                & 0.996 (0.001) & -7.36E-4 (0.001) & 0.094 (0.005) & 3.39E-3 (0.001) \\
        Occlusion           & 0.988 (0.001) & -1.93E-3 (0.001) & 0.095 (0.005) & 4.57E-3 (0.001) \\
        Aug Occlusion       & 0.989 (0.001) & 4.59E-4 (0.001) & 0.098 (0.005) & 1.90E-3 (0.002) \\
        Retain              & 0.989 (0.001) & -3.79E-3 (0.001) & 0.093 (0.005) & 7.70E-3 (0.001) \\
		\midrule
		Ours         & \textbf{0.981} (0.004) & \textbf{1.53E-2} (0.004) & \textbf{0.118} (0.008) & \textbf{-1.19E-2} (0.004) \\
		\bottomrule
	\end{tabular}%
	}
	\caption{Results of each explanation method compared with ours, by masking 20\% of the data and replacing masked features with an average over time: $\overline{x}_{t,i} = \frac{1}{T}\sum_{t} x_{t, i}$. For each metric, $\uparrow$ indicates that higher is better, and $\downarrow$ that lower is better. Mean and std are reported over 5 folds.}
	\label{tab:results_mimic_avg}
\end{table}

\begin{table}[t]
	\centering
	\resizebox{0.48\textwidth}{!}{%
	\begin{tabular}{lcccc}
		\toprule
		\textbf{Method} & Acc $\downarrow$ & Comp $\uparrow$ & CE $\uparrow$ & Suff $\downarrow$  \\
		\midrule
        DeepLift            & 0.972 (0.003) & -1.19E-3 (0.007) & 0.125 (0.014) & -6.92E-3 (0.006)  \\
        DynaMask            & 0.975 (0.002) & -1.27E-3 (0.004) & 0.106 (0.009) & 6.57E-3 (0.012)  \\
		IG                  & 0.972 (0.003) & 1.24E-4 (0.007) & 0.127 (0.015) & -7.61E-3 (0.006)  \\
        GradientShap        & 0.968 (0.006) & -6.28E-3 (0.004) & 0.128 (0.017) & 6.61E-4 (0.005)  \\
        Lime                & 0.983 (0.003) & -5.22E-3 (0.004) & 0.093 (0.008) & -2.23E-3 (0.019)  \\
        Occlusion           & 0.971 (0.003) & -4.03E-3 (0.003) & 0.122 (0.008) & -4.97E-3 (0.008)  \\
        Aug Occlusion       & 0.972 (0.003) & -6.88E-4 (0.004) & 0.121 (0.009) & -4.62E-3 (0.011)  \\
        Retain              & 0.971 (0.003) & -8.01E-3 (0.006) & 0.0123 (0.009) & 4.90E-4 (0.007)  \\
		\midrule
		Ours         & \textbf{0.943} (0.008) & \textbf{1.09E-1} (0.023) & \textbf{0.318} (0.057) & \textbf{-6.94E-2} (0.006)  \\
		\bottomrule
	\end{tabular}%
	}
	\caption{Results of each explanation method compared with ours, by masking 20\% of the data and replacing masked features with zeros: $\overline{x}_{t,i} = 0$. For each metric, $\uparrow$ indicates that higher is better, and $\downarrow$ that lower is better. Mean and std are reported over 5 folds.}
	\label{tab:results_mimic_zero}
\end{table}

\begin{figure}[t]
\begin{center}
\centerline{\includegraphics[width=\columnwidth]{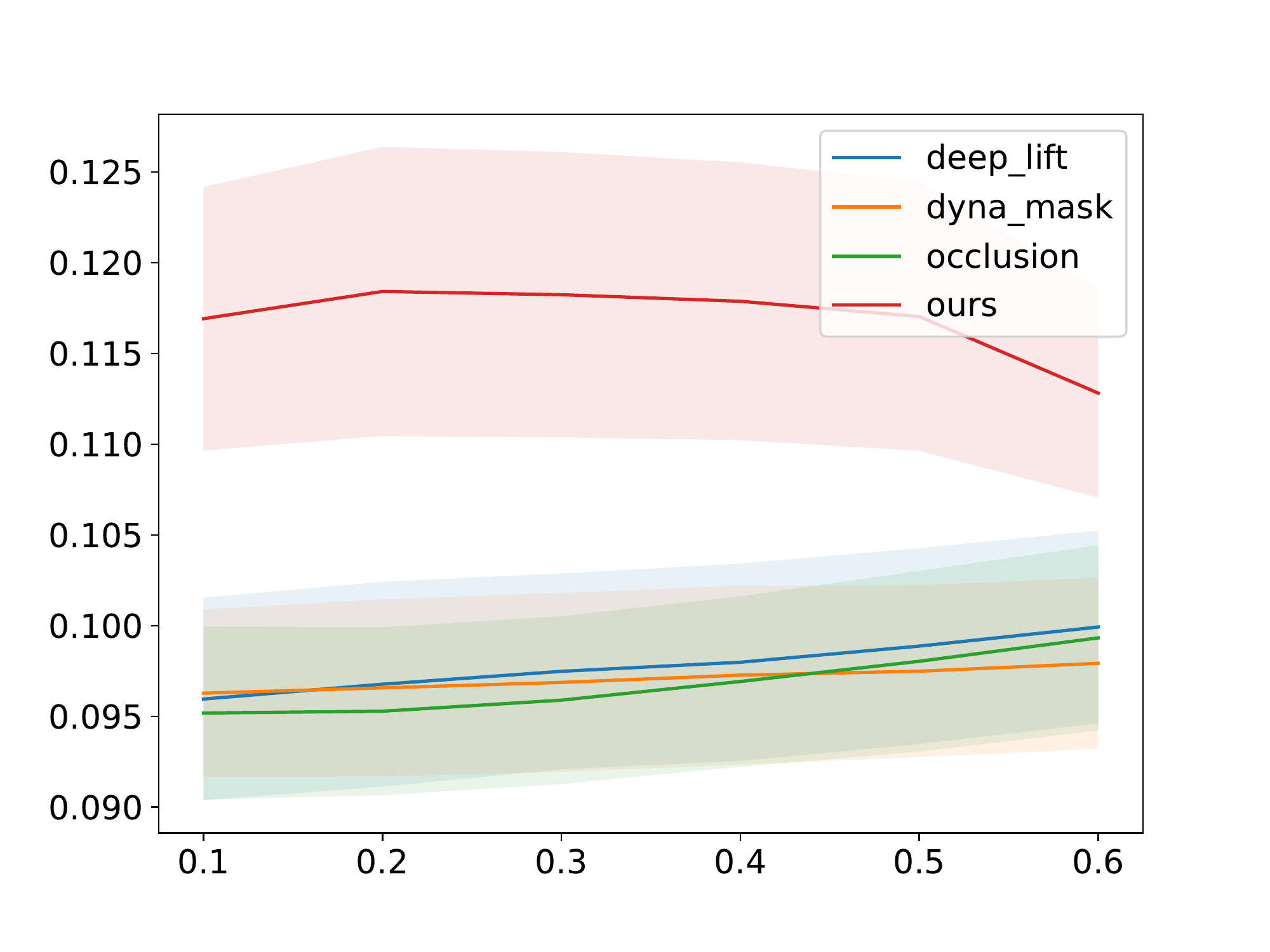}}
\caption{\textbf{Cross Entropy replacing masked data with an average.} We present here the results in terms of cross-entropy by masking between 10\% and 60\% of the data for each patient, and replacing the masked data with the overall average over time for each feature: $\overline{x}_{t,i} = \frac{1}{T}\sum_{t} x_{t, i}$. For clarity, we only plot a subset of the baselines. Higher is better with this metric.}
\label{fig:ce_average}
\end{center}
\end{figure}

\begin{figure}[t]
\begin{center}
\centerline{\includegraphics[width=\columnwidth]{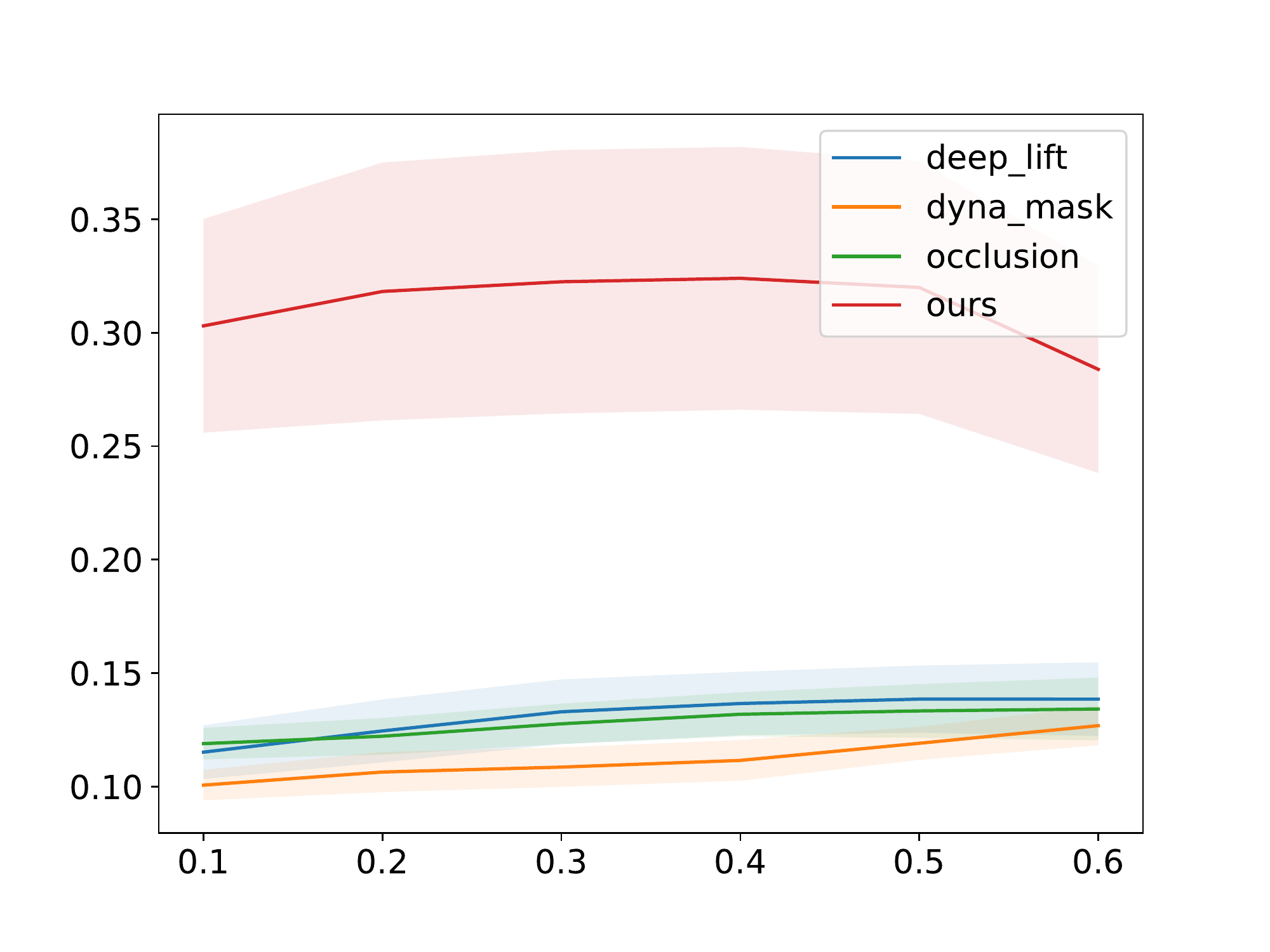}}
\caption{\textbf{Cross Entropy replacing masked data with zeros.} We present here the results in terms of cross-entropy by masking between 10\% and 60\% of the data for each patient, and replacing the masked data with zeros: $\overline{x}_{t,i} = 0$. For clarity, we only plot a subset of the baselines. Higher is better with this metric.}
\label{fig:ce_zeros}
\end{center}
\end{figure}

We present on Tables \ref{tab:results_mimic_avg} and \ref{tab:results_mimic_zero} results with our method compared with different baselines, computing our metrics by masking 20\% of the data, and replacing these features with either an average over time (Table \ref{tab:results_mimic_avg}) or zeros (Table \ref{tab:results_mimic_zero}). We also plot on Figures \ref{fig:ce_average} and \ref{fig:ce_zeros} the cross-entropy (CE) metrics by masking different proportion of the data, and replacing masked data with either an average over time (Figure \ref{fig:ce_average}) or zeros (Figure \ref{fig:ce_zeros}). We also perform ablation studies in Appendix \ref{app:lambda} and provide more results in Appendix \ref{app:results}.

Our results show that our method significantly outperforms every other method on every metric, both using the average over time or zeros as masked data. This also indicates that using learned perturbations is preferable to using fixed ones when explaining predictions on multivariate time series data.

\paragraph{Choice of the perturbation generator.} While our method seems to perform well compared with existing baselines, we want here to study the impact of the choice of NN in Equation \ref{eq:method}. In this study, we propose the following models: 

\begin{itemize}
    \item \textbf{Zero}: NN(\textbf{x}) is set to zero everywhere. Equation \ref{eq:method} is then simply: $\Phi(\textbf{x}, \textbf{m}) = \textbf{m} \times \textbf{x}$.
    \item \textbf{GRU}: We use a one layer GRU model: $\textrm{NN}(\textbf{x}) = \textrm{GRU}(\textbf{x})$, which corresponds to a generalisation of the fixed perturbation $\mu^p_{t,i}$ in \citet{crabbe2021explaining}.
    \item \textbf{Bi-GRU}: Finally, we use a one layer bidirectional GRU $\textrm{NN}(\textbf{x}) = \textrm{bi-GRU}(\textbf{x})$, which corresponds to a generalisation of the fixed perturbation $\mu_{t,i}$ in \citet{crabbe2021explaining}.
\end{itemize}

\begin{table}[t]
	\centering
	\resizebox{0.48\textwidth}{!}{%
	\begin{tabular}{lcccc}
		\toprule
		\textbf{Method} & Acc $\downarrow$ & Comp $\uparrow$ & CE $\uparrow$ & Suff $\downarrow$  \\
		\midrule
        Zeros               & 0.981 (0.003) & 1.36E-2 (0.001) & 0.116 (0.004) & -1.02E-2 (0.002)  \\
		GRU                 & \textbf{0.980} (0.004) & \textbf{1.76E-2} (0.001) & \textbf{0.122} (0.004) & \textbf{-1.37E-2} (0.002)  \\
        Bi-GRU              & 0.981 (0.004) & 1.53E-2 (0.004) & 0.118 (0.008) & -1.19E-2 (0.004)  \\
		\bottomrule
	\end{tabular}%
	}
	\caption{Comparison of different perturbation models, masking 20\% of the data and replacing masked features with an average over time: $\overline{x}_{t,i} = \frac{1}{T}\sum_{t} x_{t, i}$. For each metric, $\uparrow$ indicates that higher is better, and $\downarrow$ that lower is better. Mean and std are reported over 5 folds.}
	\label{tab:ablation_mimic_avg}
\end{table}

\begin{table}[t]
	\centering
	\resizebox{0.48\textwidth}{!}{%
	\begin{tabular}{lcccc}
		\toprule
		\textbf{Method} & Acc $\downarrow$ & CE $\uparrow$ & Comp $\uparrow$ & Suff $\downarrow$  \\
		\midrule
        Zeros               & 0.951 (0.005) & 9.64E-2 (0.013) & 0.305 (0.015) & -6.79E-2 (0.002)  \\
		GRU                 & \textbf{0.943} (0.007) & \textbf{1.22E-1} (0.008) & \textbf{0.344} (0.017) & \textbf{-7.40E-2} (0.001)  \\
        Bi-GRU              & \textbf{0.943} (0.008) & 1.09E-1 (0.023) & 0.318 (0.057) & -6.94E-2 (0.006)  \\
		\bottomrule
	\end{tabular}%
	}
	\caption{Comparison of different perturbation models, masking 20\% of the data and replacing masked features with zeros: $\overline{x}_{t,i} = 0$. For each metric, $\uparrow$ indicates that higher is better, and $\downarrow$ that lower is better. Mean and std are reported over 5 folds.}
	\label{tab:ablation_mimic_zero}
\end{table}

We present our results on MIMIC-III on Tables \ref{tab:ablation_mimic_avg} and \ref{tab:ablation_mimic_zero}, replacing 20\% of the data with either an overall average of each feature over time (Table \ref{tab:ablation_mimic_avg}), or with zeros (Table \ref{tab:ablation_mimic_zero}). We use the same metrics as with our main MIMIC-III experiments. As with the main experiment, we provide more results, masking different proportions of the data, in Appendix \ref{app:results}.

Our results are interesting on several accounts. Firstly, the Zeros method, which simply perturbs the data by masking non salient features: $\Phi(\textbf{x}, \textbf{m}) = \textbf{m} \times \textbf{x}$, performs significantly better than all other baselines, including DynaMask with fixed perturbations. As each measure in our dataset is normalised, masking one measure with the Zeros method amounts to replacing it with its average over the entire dataset. On the other hand, DynaMask replaced mask data with its average over time \textit{for each individual patient}. This good performance of Zeros could be therefore explained by the fact that many measures do not vary much over time. As a result, replacing masked data with an overall average would be much more informative than replacing it with an average over time for each patient.

Secondly, while using the bidirectional GRU perturbation yields better results than Zeros, it is itself outperformed by our method with the unidirectional GRU perturbation. Moreover, using this unidirectional GRU also yields more stable results with a lower standard deviation. Our intuition was that a bidirectional GRU would yield better results, as it would be able to produce outputs based on past and future events. However, it seems that modeling perturbations ignoring future events seems to yield better and more stable results. We used a Bi-GRU to produce our results in Tables \ref{tab:results_mimic_avg} and \ref{tab:results_mimic_zero}, as it corresponds to our original intuition, but we also recommend testing different types of neural networks for best performance when applying our method.

\begin{figure}[t]
\begin{center}
\centerline{\includegraphics[width=\columnwidth]{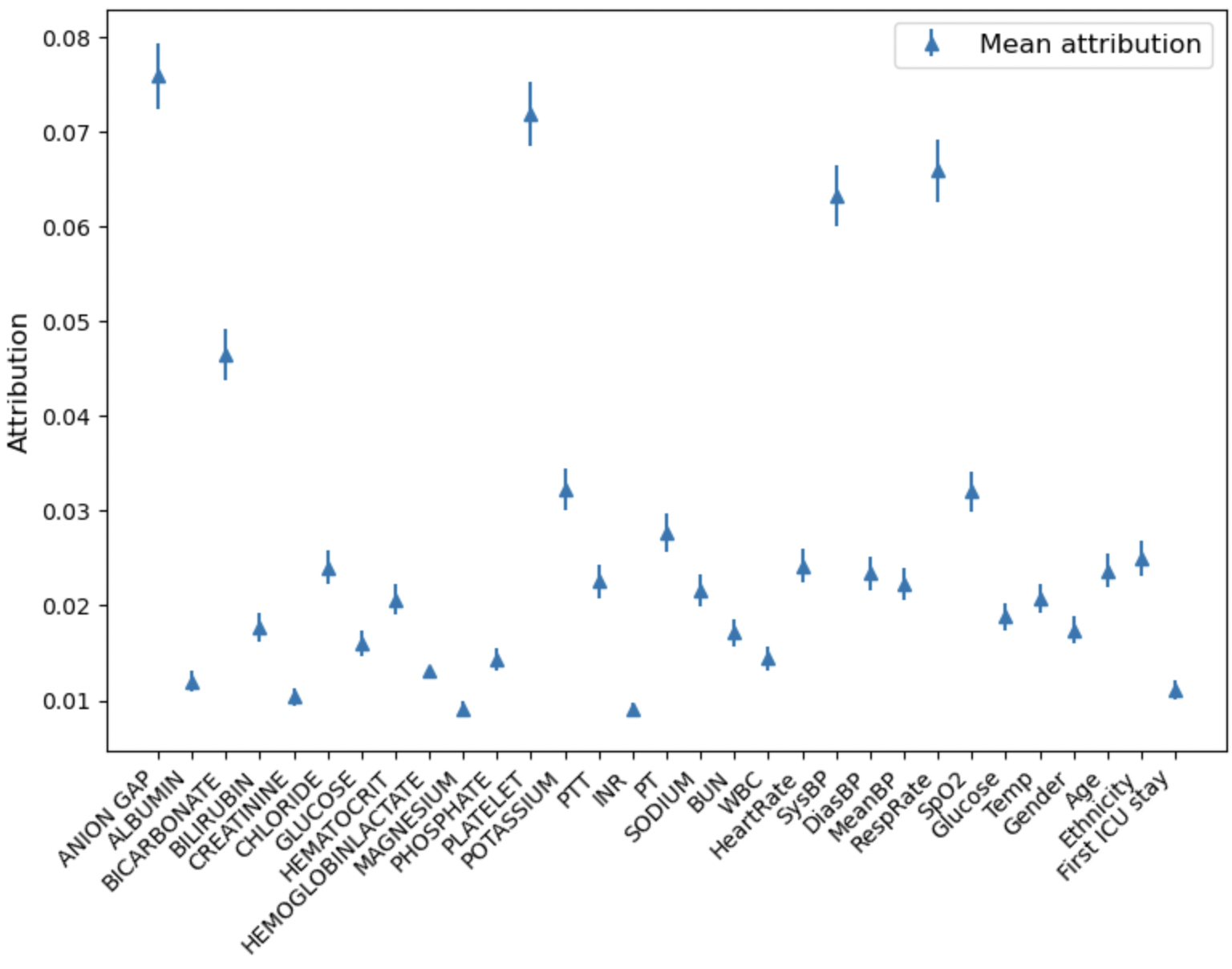}}
\caption{\textbf{Importance of each feature to predict in-hospital mortality.} For each feature, we present its average importance over time and over multiple patients, using our method with a GRU perturbation network. We infer from these results that anion gap, bicarbonate level, platelet count, systolic blood pressure and respiratory rate are most important for our model when making a prediction. We also plot a 95\% confidence interval around these averages.}
\label{fig:features_imp}
\end{center}
\vskip -0.1in
\end{figure}

\paragraph{Analysis of salient features.} We present on Figure \ref{fig:features_imp} the most salient features, averaged over every positive patient, to determine which factors are most important when determining in-hospital mortality.

This averaged feature importance indicates a few salient features: anion gap, bicarbonate level, platelet count, systolic blood pressure and respiratory rate. This seems to be consistent with the literature, which has highlighted the importance of these features, conducting studies on the saliency of bicarbonate levels \citep{lim2019short}, platelet count \citep{zhang2014platelet} and systolic blood pressure \citep{kondo2011revised}. The influence of anion gap on in-hospital mortality is less clear, with conflicting studies on this subject \citep{glasmacher2015anion}. On the other hand, the respiratory rate is often neglected despite being an important predictor of serious events \cite{cretikos2008respiratory}.

However, although the 95\% confidence interval associated with these features importances is small due to a large number of patients, there remains a large corresponding standard deviation. We can therefore infer that the importance of each feature greatly depends on each patient. It is indeed possible that a measure such as the systolic blood pressure, for instance, only matters when it is outside of a normal range. As a result, its importance will greatly vary depending on each patient's condition. This demonstrates the superiority of perturbation-based methods compared to directly using a simpler interpretable model such as a decision tree instead of a neural network to predict in-hospital mortality. Indeed, such methods can only infer feature importance on average, and not explain each prediction individually.

\begin{figure}[t]
\begin{center}
\centerline{\includegraphics[width=\columnwidth]{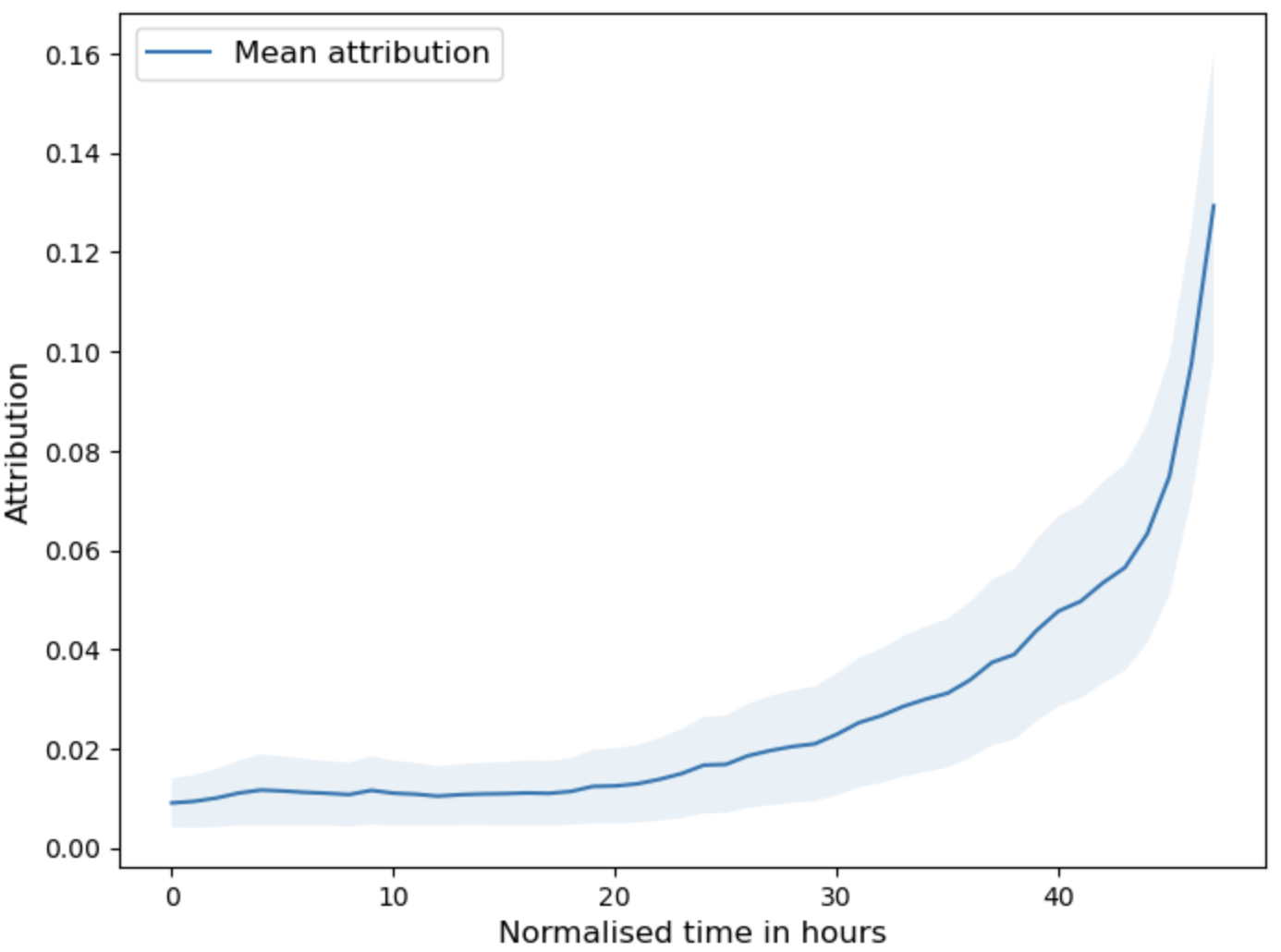}}
\caption{\textbf{Average feature importance over time to predict in-hospital mortality.} We average every feature's importance on all positive patients, over each of the 48 hours of measurement in hospital. We display the mean with a 95\% confidence interval. Our results show that later measures are most important predictor of in-hospital mortality.}
\label{fig:temp_imp}
\end{center}
\vskip -0.1in
\end{figure}

\begin{figure}[t]
\begin{center}
\centerline{\includegraphics[width=\columnwidth]{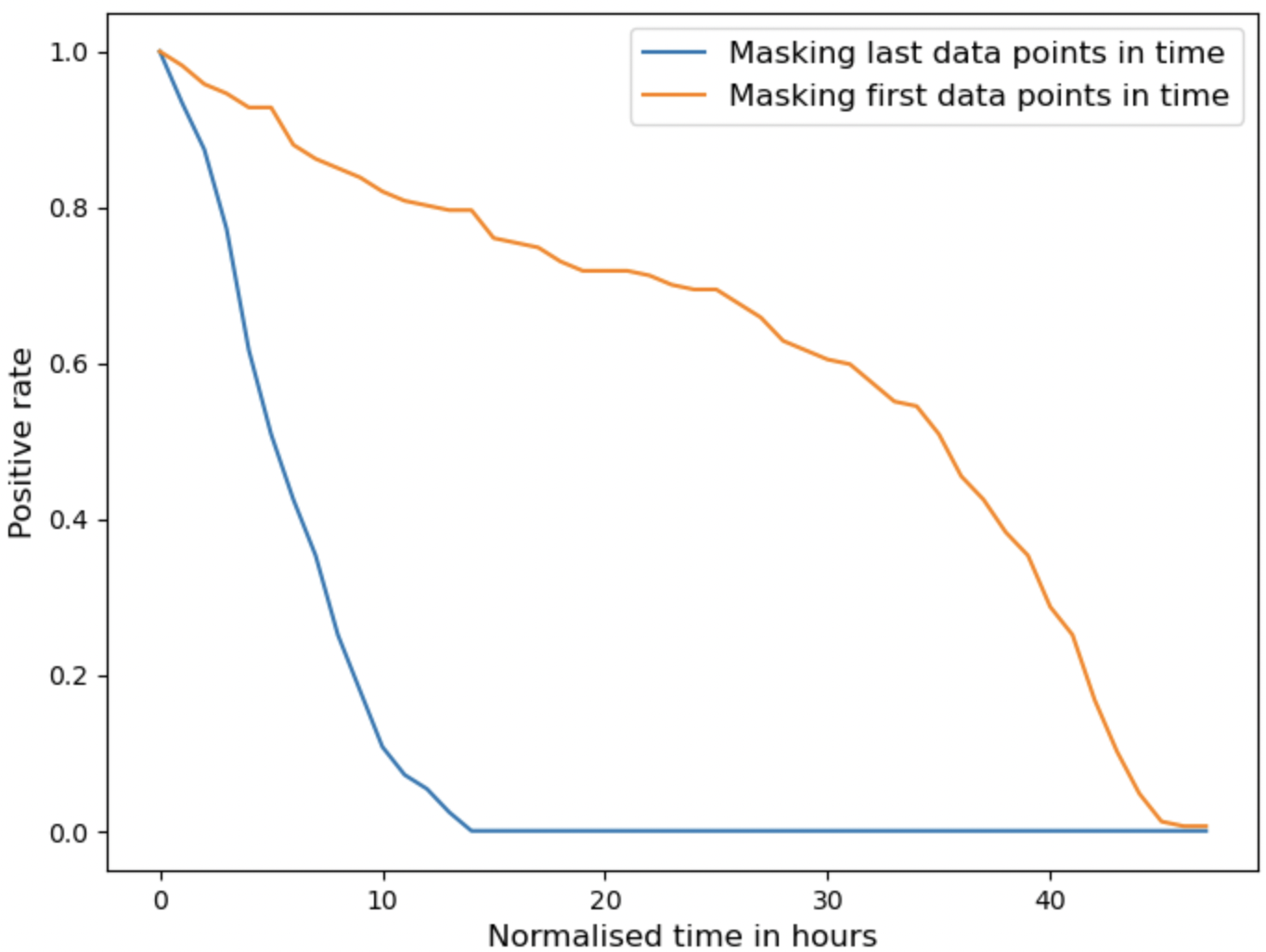}}
\caption{\textbf{Positive rate over positive patients by masking first or last data points.} We compare the influence of the first or last data points in time by masking them successively. We observe that masking latter points in time yields a much lower positive rate than masking former ones. Masking the last 12 measures indeed results in every positive patient predicted as negative.}
\label{fig:masking_points}
\end{center}
\vskip -0.1in
\end{figure}

In addition to determining which feature is salient, our method can also infer \textbf{when} it is salient. As a result, we present on Figure \ref{fig:temp_imp} the average over positive patients of the importance of all features at each hour. This figure shows that later measurements have a larger impact on the outcome compared with earlier data. To evaluate the accuracy of this finding, we also plot on Figure \ref{fig:masking_points} the positive rate over (true or false) positive patients, when masking earlier measures on one hand, and later measures on the other hand. We can see that masking early features has a minimal impact on the predictions, while masking late features has on the contrary a dramatic impact on the outcome. As a result, it seems that, when predicting in-hospital mortality, the last measurements of each patient is more important to make a prediction, rather than the overall evolution of the patient.

\section{Conclusion}
\label{conclusion}

In this work, we have presented an extension of \citet{fong2017interpretable} and \citet{crabbe2021explaining} to better explain multivariate time series predictions using a perturbation-based saliency method. Our main intuition is that the choice made by \citet{crabbe2021explaining} of fixed perturbations is less adapted to temporal data due to the possibility of long-term dependencies. 

Our results show that using learned perturbations yields better explanations compared with existing methods, including the DynaMask one with fixed perturbations. We have also studied the choice of the neural network to model the perturbation and found that, on the in-hospital mortality task of MIMIC-III, a unidirectional GRU yields better and more stable results than the bidirectional one.

Using our method, we have also been able to derive some insights on the neural network predicting in-hospital mortality: which features are on average most important, as well as which measures in time. Precise temporal attributions could be derived similarly for each patient, giving further insight on this model's behavior.

Moreover, an inherent limitation to perturbation-based methods such as ours is that it is not able to specify the direction of an explanation. As such, it can measure if a specific feature is important, but cannot distinguish between features having a positive or a negative influence on the prediction. Adapting our method to tackle this issue would prove very beneficial for applications in healthcare.

\section{Acknowledgements}
\label{acknowledgements}

The author would like to thank Vitalii Zhelezniak for his insightful remarks and recommendations during the elaboration of this work. We also thank Anthony Hu and Thomas Uriot for their detailed initial reviews of this paper.


\bibliography{bib}
\bibliographystyle{icml2023}

\newpage
\appendix
\onecolumn
\section{Addition studies using the Mimic3 dataset}
\label{app:lambda}

First, we perform here an ablation study to determine which values of $\lambda_1$ and $\lambda_2$ should be used in Equation~\ref{eq:optim_modif} on the Mimic3 dataset, similarly to the one done on HMM in Section \ref{experiments}. We run our experiment using various values of $\lambda_1$ and $\lambda_2$, running on 5 different seed for each pair of parameters. We report our results on Table \ref{tab:ablation_lambda_mimic}.

This table shows similarly that a high value of $\lambda_2$ should be chosen, to force NN(x) to learn uninformative perturbations. Interestingly, using a low value of $\lambda_1$ yields stronger results on Mimic3, indicating that constraining too much $\textbf{m}$ to be close to $\textbf{0}$ is not necessarily a good option. However, this setting does not yield good results on the HMM dataset, therefore our method should be carefully evaluated when used on low values of $\lambda_1$.

\begin{table}[ht]
	\centering
	\begin{tabular}{lc|ccccc}
		\toprule
        \multicolumn{2}{c}{} & \multicolumn{5}{c}{$\lambda_1$} \\
        \multicolumn{2}{c}{} & 0.01 & 0.1 & 1 & 10 & 100 \\
        \cmidrule{3-7}
        \multirow{5}{*}{$\lambda_2$} & 
        0.01 & 0.926 - 0.348 & 0.965 - 0.160 & 0.968 - 0.186 & 0.968 - 0.165 & 0.956 - 0.169 \\
        & 0.1 & 0.893 - 0.503 & 0.935 - 0.328 & 0.961 - 0.166 & 0.961 - 0.261 & 0.936 - 0.271 \\
        & 1 & 0.881- 0.534 & 0.899 - 0.491 & 0.947 - 0.284 & 0.957 - 0.191 & 0.960 - 0.261 \\
        & 10 & 0.857 - 0.540 & 0.905 - 0.480 & 0.940 - 0.372 & 0.950 - 0.266 & 0.964 - 0.163 \\
        & 100 & 0.862 - 0.542 & 0.910 - 0.472 & 0.950 - 0.362 & 0.946 - 0.289 & 0.956 - 0.173 \\
		\bottomrule
	\end{tabular}%
	\caption{Influence of $\lambda_1$ and $\lambda_2$ from Equation \ref{eq:optim_modif} on the results of the Mimic3 experiment. or each pair of parameters, 2 values are reported: Accuracy - Cross-entropy. For accuracy, lower is better, while higher is better for cross-entropy. Each metric is computed by masking 20 \% of the data and replacing masked features with zeros: $\overline{x}_{t,i} = 0$. The average result over 5 runs is reported.}
	\label{tab:ablation_lambda_mimic}
\end{table}

\vspace{1cm}

Second, we learn perturbations as a deletion game, similarly to the experiment conducted on the HMM dataset in Section \ref{experiments}. As such, we use Equation \ref{eq:deletion_modif} in this experiment. We report our results on Table \ref{tab:loss_study_mimic}.

Similarly to the HMM experiment, this table shows that the deletion mode performs poorly compared with the preservation one. The latter mode should therefore be preferred to the former.

\begin{table}[ht]
	\centering
	\begin{tabular}{lcccc}
		\toprule
		\textbf{Mode} & Acc $\downarrow$ & Comp $\uparrow$ & CE $\uparrow$ & Suff $\downarrow$ \\
		\midrule
		Preservation         & \textbf{0.943} (0.008) & \textbf{1.09E-1} (0.023) & \textbf{0.318} (0.057) & \textbf{-6.94E-2} (0.006)  \\
        Deletion            & 0.977 (0.003) & -0.025 (0.009) & 0.079 (0.012) & 0.053 (0.013) \\      
		\bottomrule
	\end{tabular}%
	\caption{Comparison of using the preservation mode vs deletion mode on the Mimic3 experiment. The average result over 5 runs is reported.}
	\label{tab:loss_study_mimic}
\end{table}

\newpage

\section{Additional results on the in-hospital mortality experiment}
\label{app:results}

\begin{figure}[h]
    \centering
    \begin{minipage}{0.45\textwidth}
        \centering
        \includegraphics[width=0.9\textwidth]{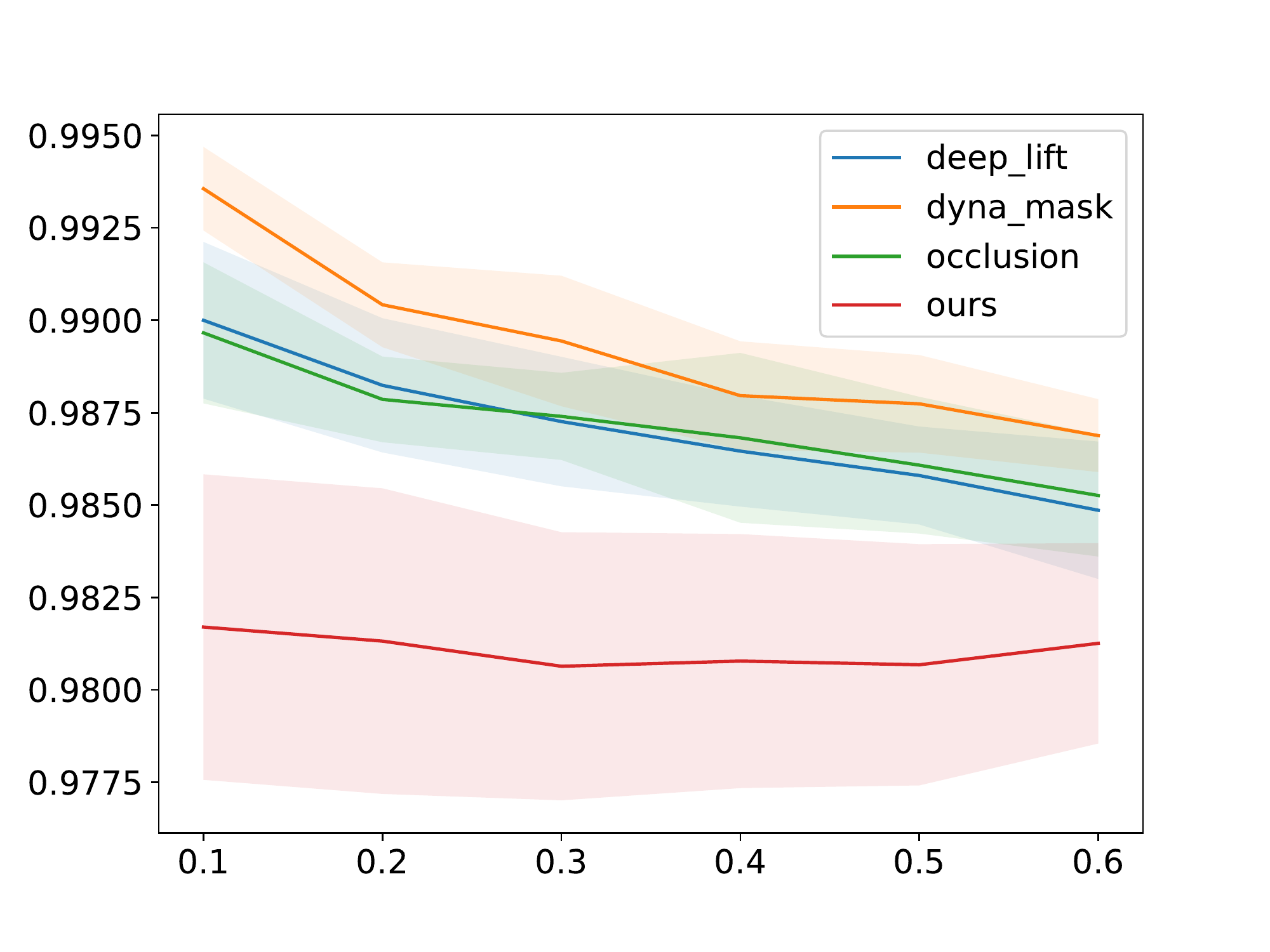} 
        \caption{Accuracy, masking between 10\% and 60\% of the data for each patient, and replacing the masked data with the overall average over time for each feature: $\overline{x}_{t,i} = \frac{1}{T}\sum_{t} x_{t, i}$. For clarity, we only plot a subset of the baselines. Lower is better with this metric.}
        \label{fig:acc_avg}
    \end{minipage}\hfill
    \begin{minipage}{0.45\textwidth}
        \centering
        \includegraphics[width=0.9\textwidth]{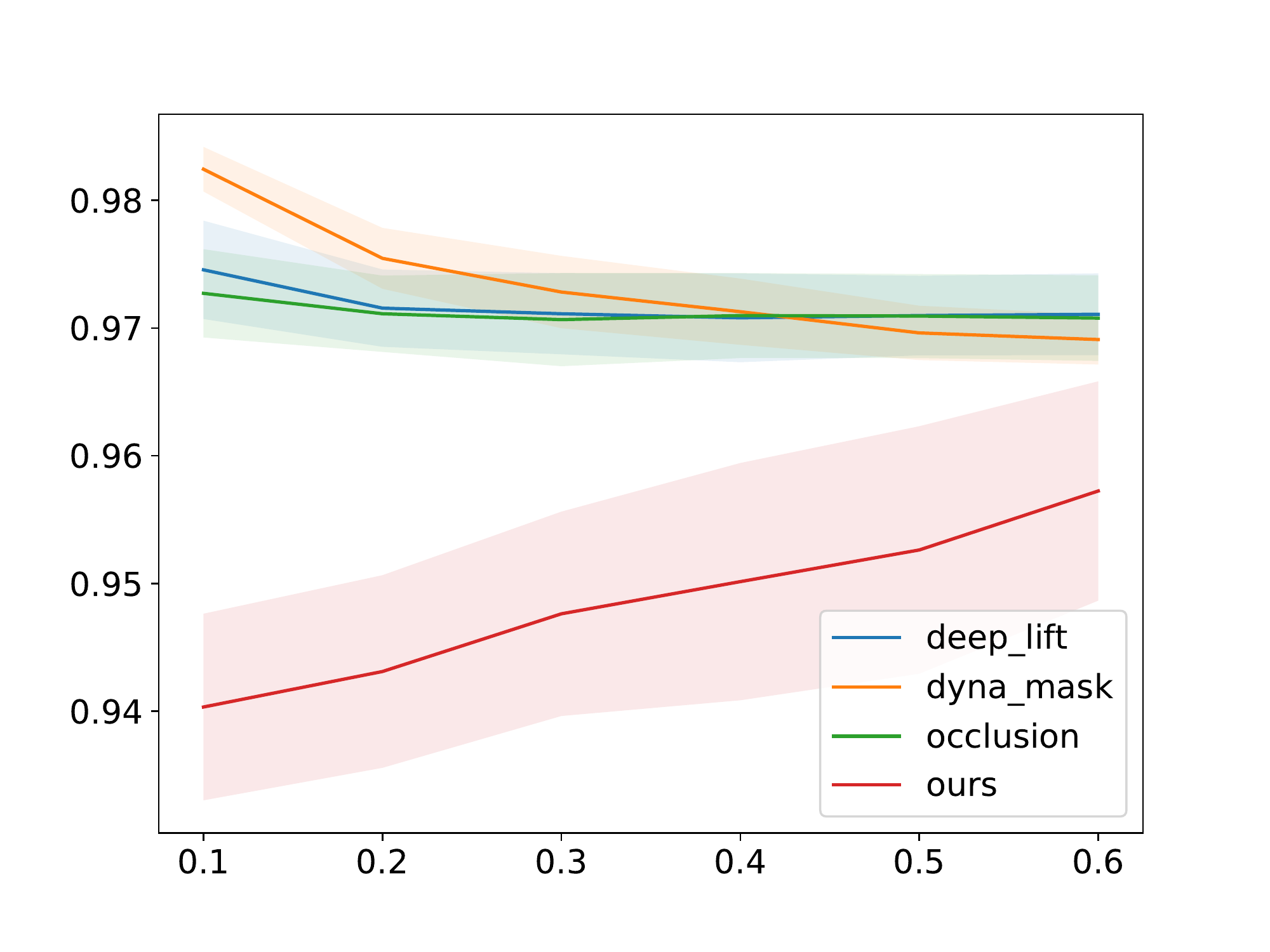} 
        \caption{Accuracy, masking between 10\% and 60\% of the data for each patient, and replacing the masked data with zeros: $\overline{x}_{t,i} = 0$. For clarity, we only plot a subset of the baselines. Lower is better with this metric.}
        \label{fig:acc_zeros}
    \end{minipage}
\end{figure}

\begin{figure}[h]
    \centering
    \begin{minipage}{0.45\textwidth}
        \centering
        \includegraphics[width=0.9\textwidth]{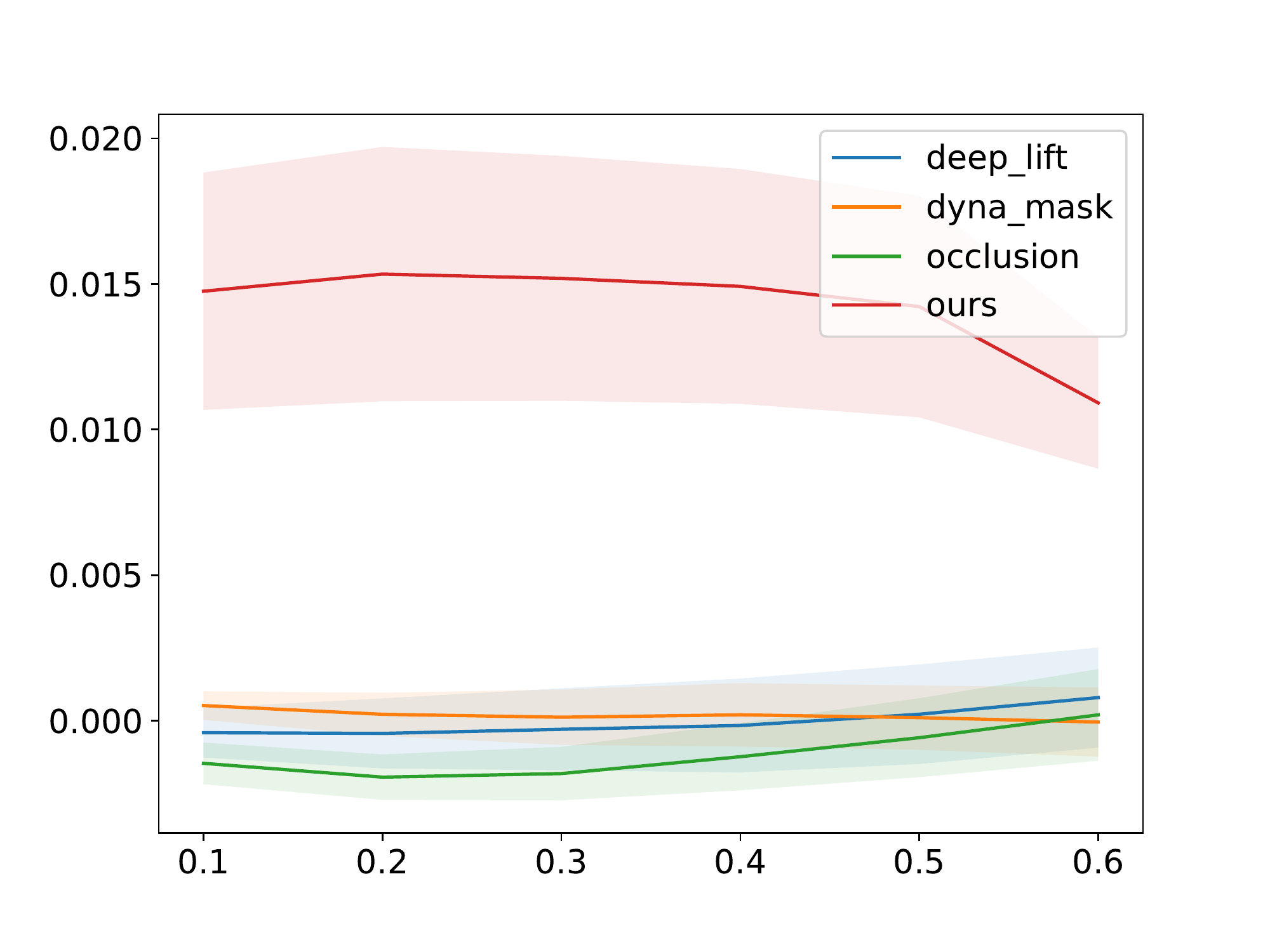} 
        \caption{Comprehensiveness, masking between 10\% and 60\% of the data for each patient, and replacing the masked data with the overall average over time for each feature: $\overline{x}_{t,i} = \frac{1}{T}\sum_{t} x_{t, i}$. For clarity, we only plot a subset of the baselines. Higher is better with this metric.}
        \label{fig:comp_avg}
    \end{minipage}\hfill
    \begin{minipage}{0.45\textwidth}
        \centering
        \includegraphics[width=0.9\textwidth]{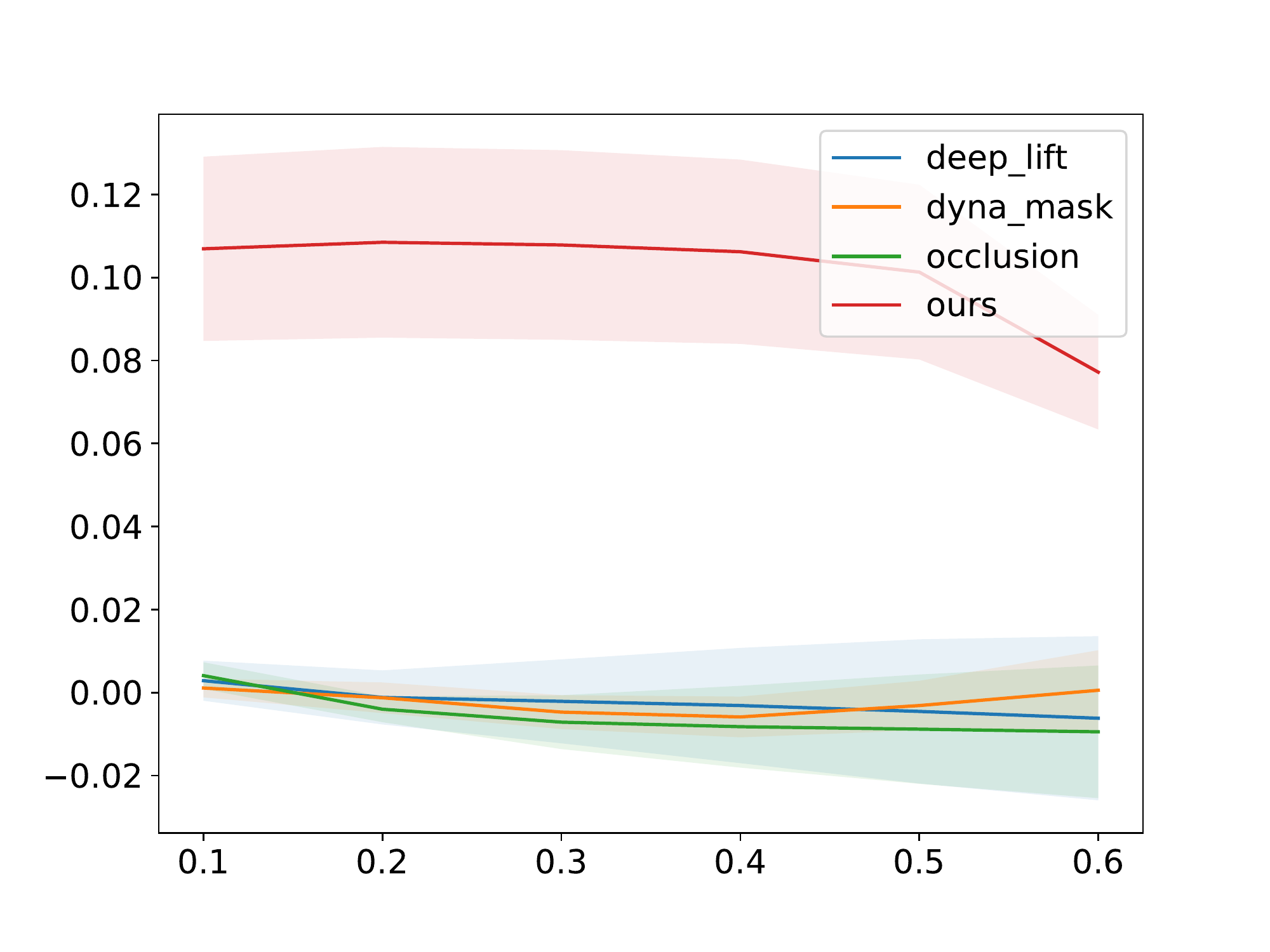} 
        \caption{Comprehensiveness, masking between 10\% and 60\% of the data for each patient, and replacing the masked data with zeros: $\overline{x}_{t,i} = 0$. For clarity, we only plot a subset of the baselines. Higher is better with this metric.}
        \label{fig:comp_zeros}
    \end{minipage}
\end{figure}

\begin{figure}[h]
    \centering
    \begin{minipage}{0.45\textwidth}
        \centering
        \includegraphics[width=0.9\textwidth]{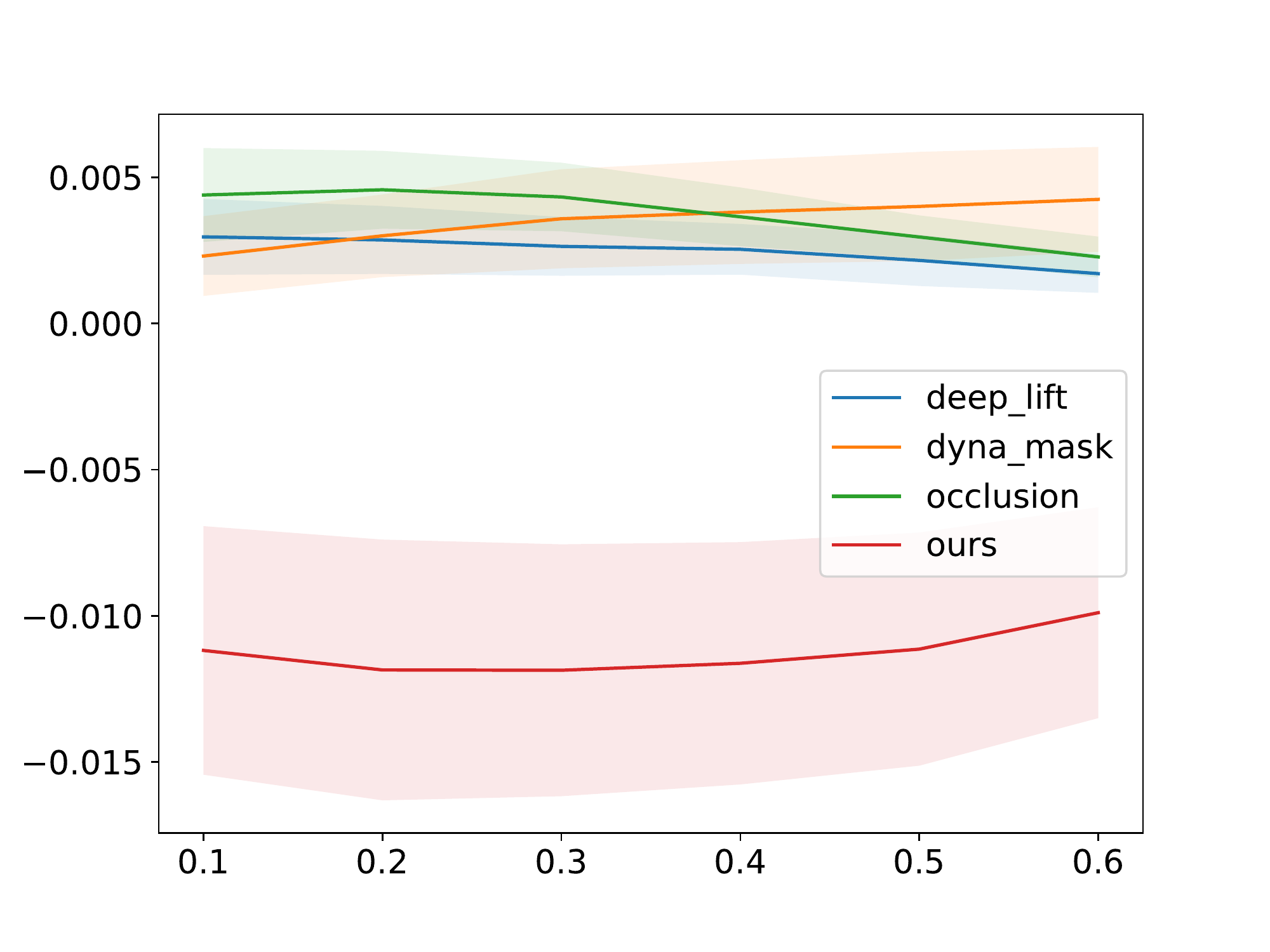} 
        \caption{Sufficiency, masking between 10\% and 60\% of the data for each patient, and replacing the masked data with the overall average over time for each feature: $\overline{x}_{t,i} = \frac{1}{T}\sum_{t} x_{t, i}$. For clarity, we only plot a subset of the baselines. Lower is better with this metric.}
        \label{fig:suff_avg}
    \end{minipage}\hfill
    \begin{minipage}{0.45\textwidth}
        \centering
        \includegraphics[width=0.9\textwidth]{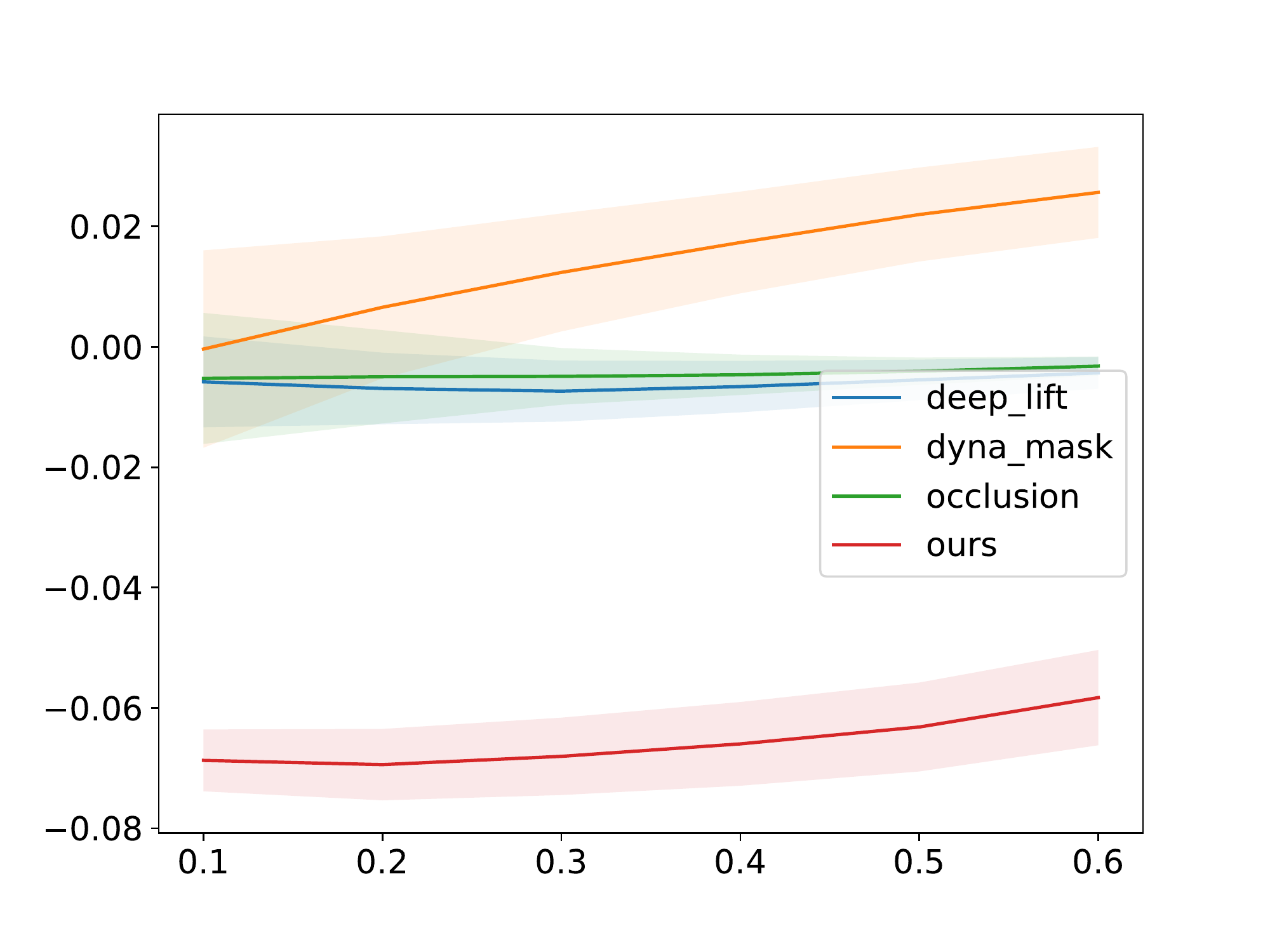} 
        \caption{Sufficiency, masking between 10\% and 60\% of the data for each patient, and replacing the masked data with zeros: $\overline{x}_{t,i} = 0$. For clarity, we only plot a subset of the baselines. Lower is better with this metric.}
        \label{fig:suff_zeros}
    \end{minipage}
\end{figure}

We present below more results on the in-hospital mortality experiment, based on the MIMIC-III dataset. Results in terms of accuracy, comprehensiveness and sufficiency can be found on Figures \ref{fig:acc_avg}, \ref{fig:acc_zeros}, \ref{fig:comp_avg}, \ref{fig:comp_zeros}, \ref{fig:suff_avg} and \ref{fig:suff_zeros}.

\begin{figure}[h]
    \centering
    \begin{minipage}{0.45\textwidth}
        \centering
        \includegraphics[width=0.9\textwidth]{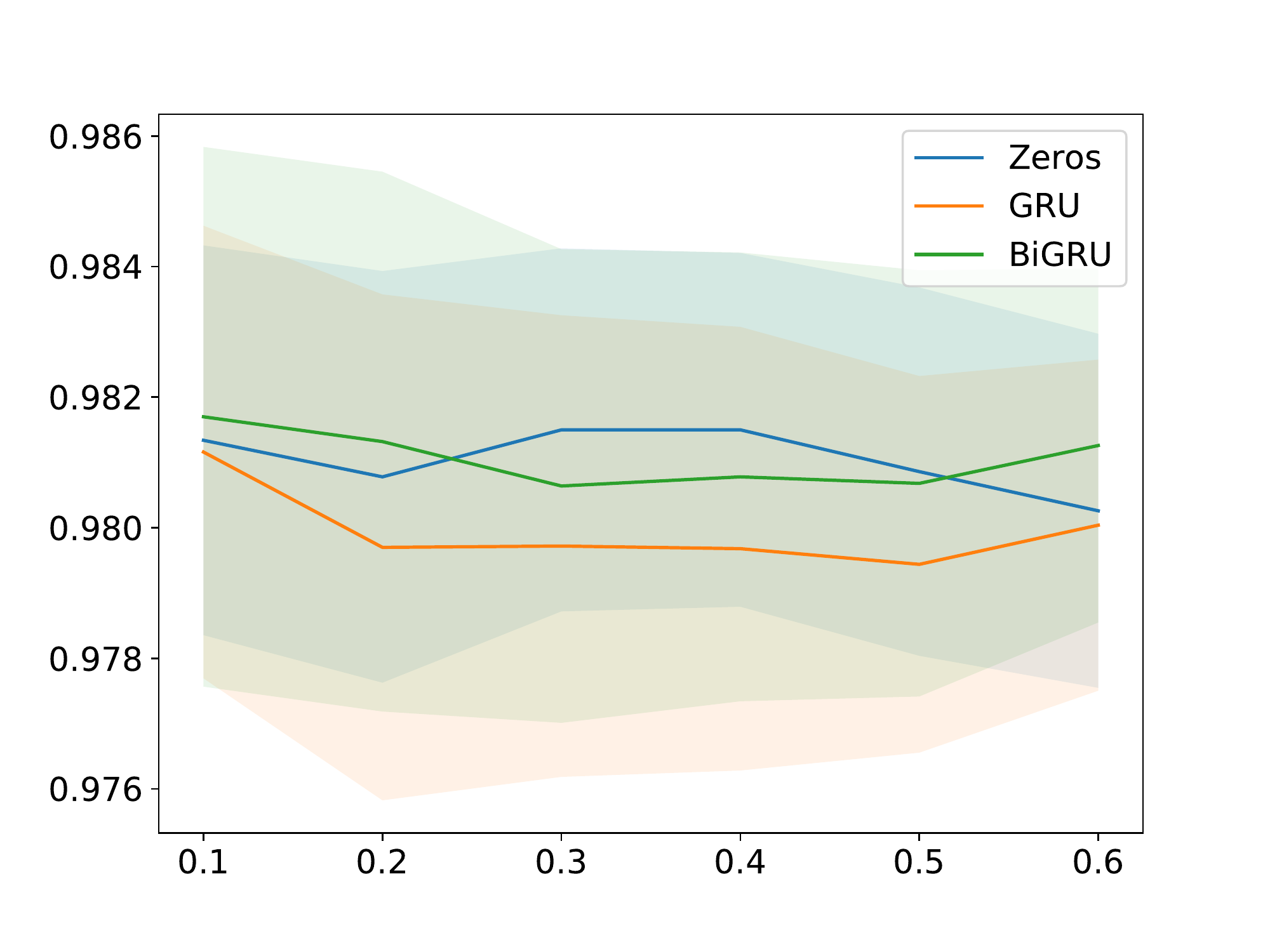} 
        \caption{Accuracy results, masking between 10\% and 60\% of the data for each patient, and replacing the masked data with the overall average over time for each feature: $\overline{x}_{t,i} = \frac{1}{T}\sum_{t} x_{t, i}$. Lower is better with this metric.}
        \label{fig:acc_avg_abl}
    \end{minipage}\hfill
    \begin{minipage}{0.45\textwidth}
        \centering
        \includegraphics[width=0.9\textwidth]{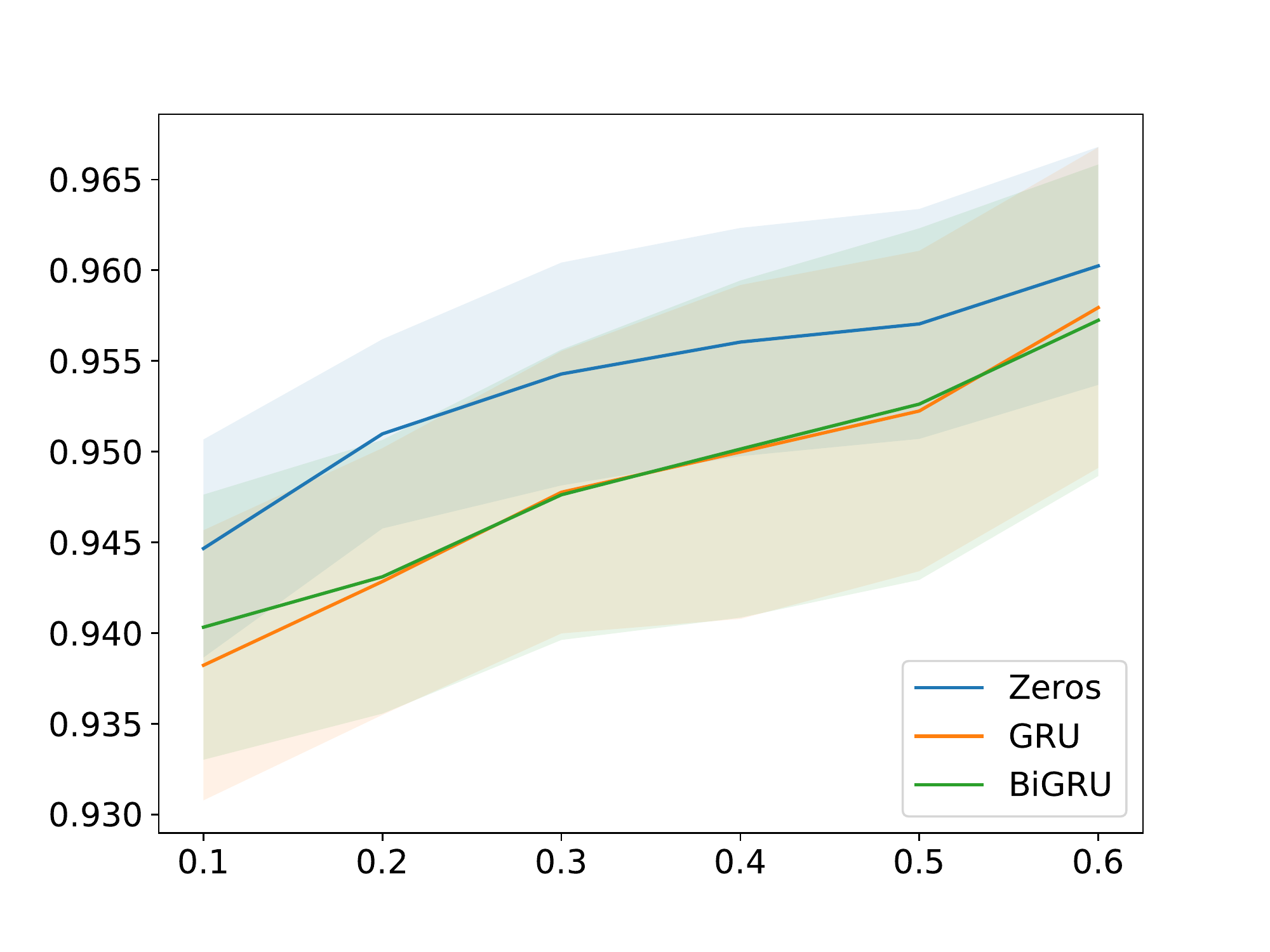} 
        \caption{Accuracy results, masking between 10\% and 60\% of the data for each patient, and replacing the masked data with zeros: $\overline{x}_{t,i} = 0$. Lower is better with this metric.}
        \label{fig:acc_zeros_abl}
    \end{minipage}
\end{figure}

\begin{figure}[h]
    \centering
    \begin{minipage}{0.45\textwidth}
        \centering
        \includegraphics[width=0.9\textwidth]{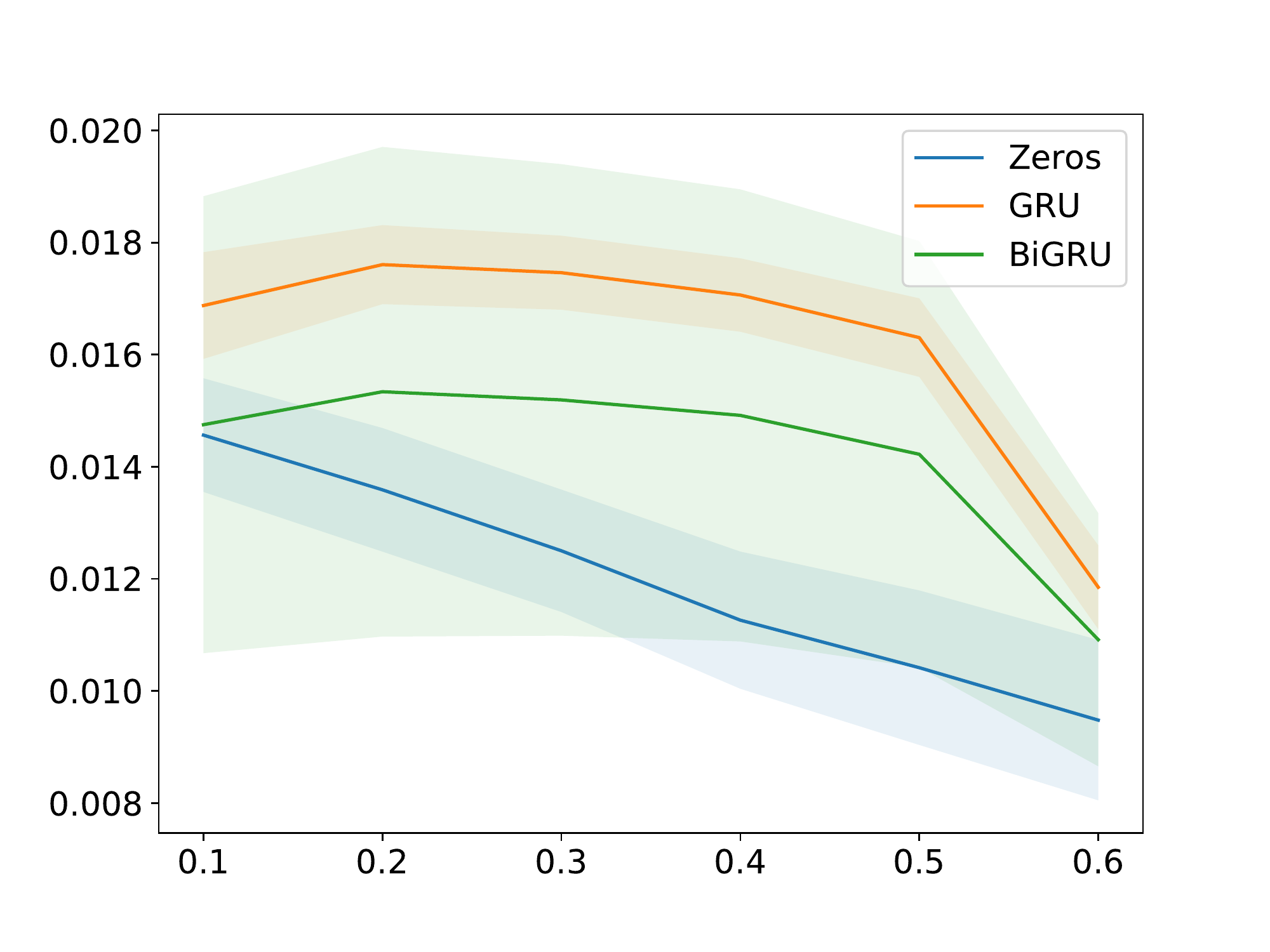} 
        \caption{Comprehensiveness results, masking between 10\% and 60\% of the data for each patient, and replacing the masked data with the overall average over time for each feature: $\overline{x}_{t,i} = \frac{1}{T}\sum_{t} x_{t, i}$. Higher is better with this metric.}
        \label{fig:comp_avg_abl}
    \end{minipage}\hfill
    \begin{minipage}{0.45\textwidth}
        \centering
        \includegraphics[width=0.9\textwidth]{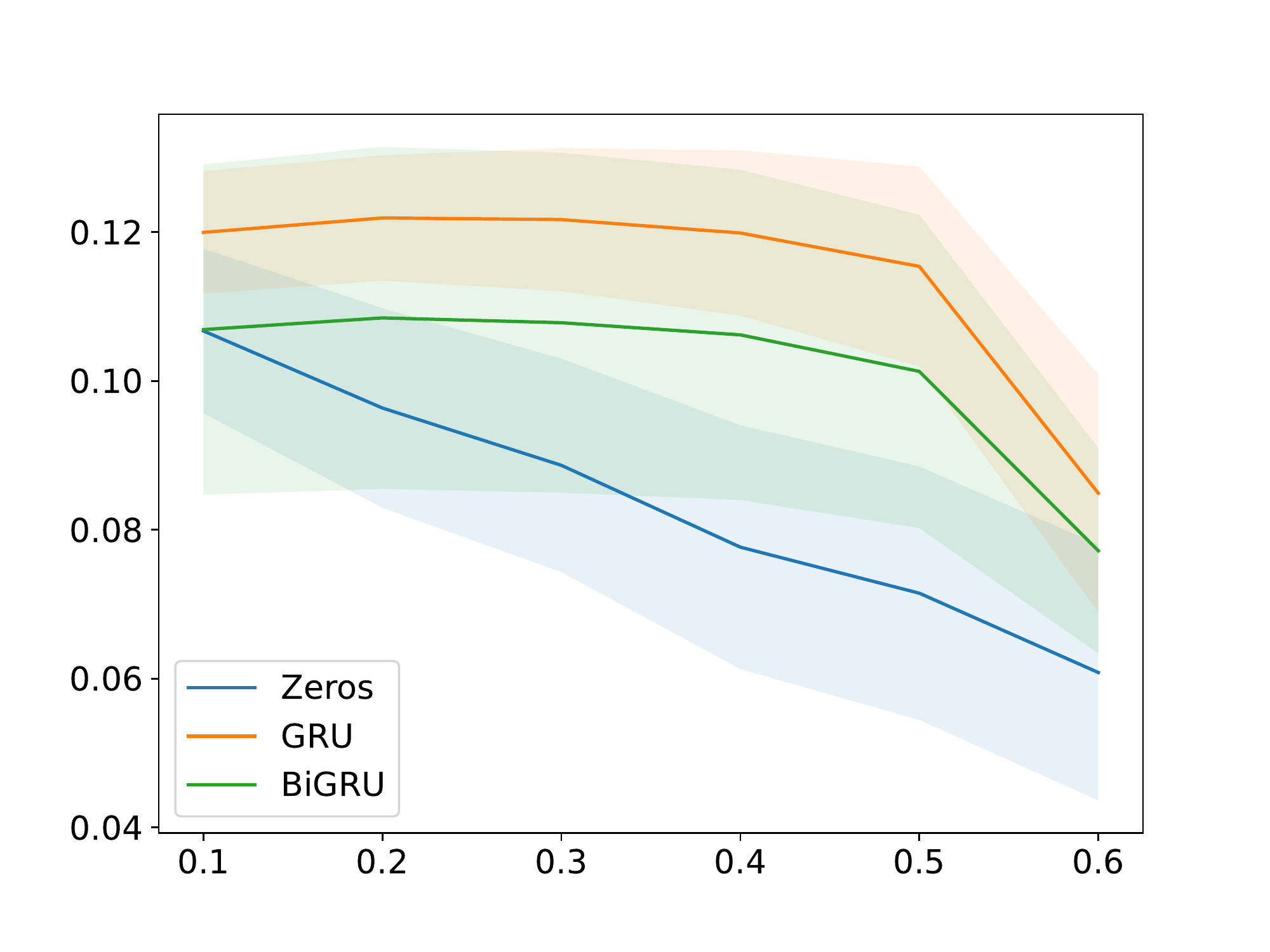} 
        \caption{Comprehensiveness results, masking between 10\% and 60\% of the data for each patient, and replacing the masked data with zeros: $\overline{x}_{t,i} = 0$. Higher is better with this metric.}
        \label{fig:comp_zeros_abl}
    \end{minipage}
\end{figure}

\begin{figure}[h]
    \centering
    \begin{minipage}{0.45\textwidth}
        \centering
        \includegraphics[width=0.9\textwidth]{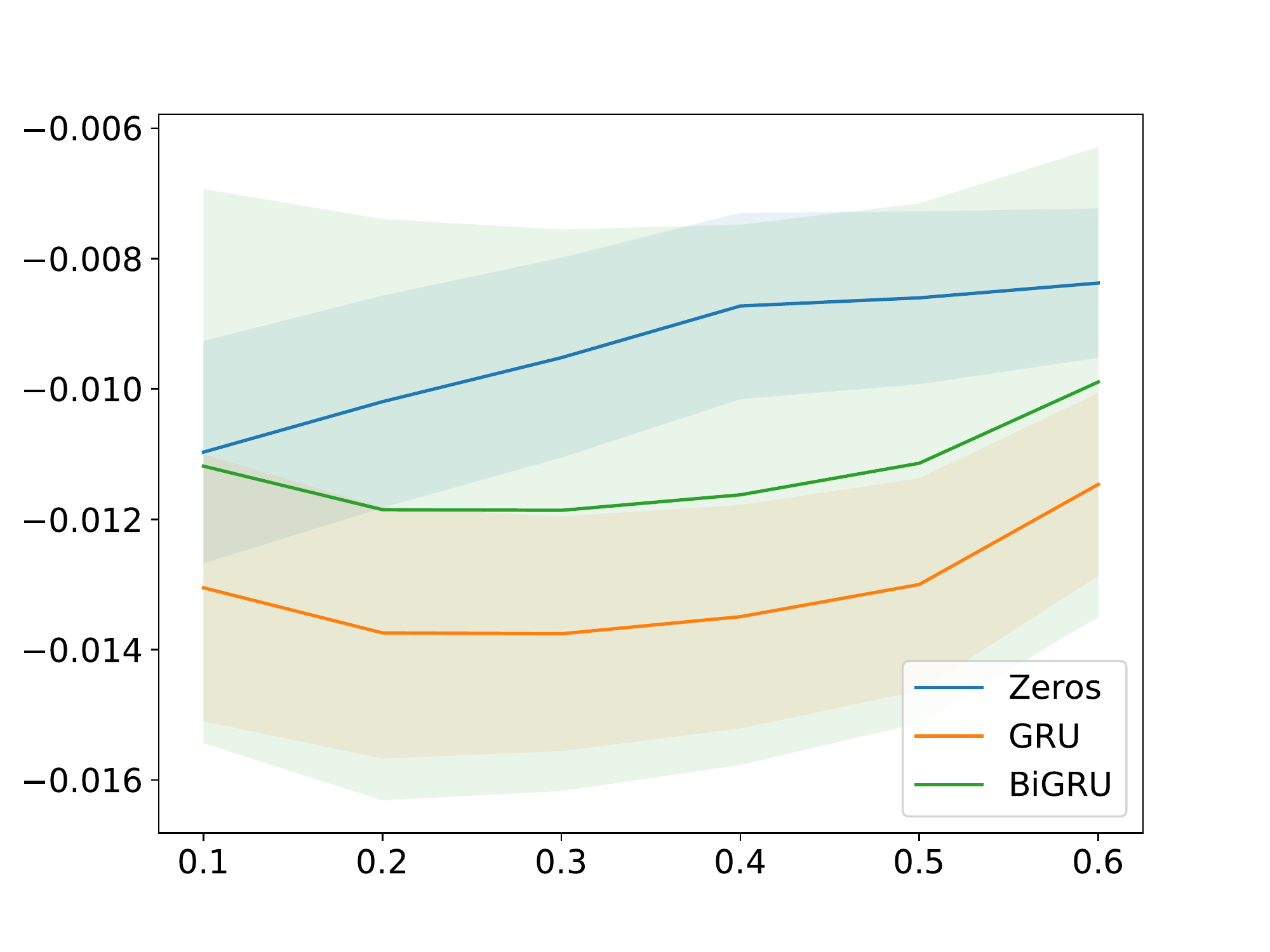} 
        \caption{Sufficiency results, masking between 10\% and 60\% of the data for each patient, and replacing the masked data with the overall average over time for each feature: $\overline{x}_{t,i} = \frac{1}{T}\sum_{t} x_{t, i}$. Lower is better with this metric.}
        \label{fig:suff_avg_abl}
    \end{minipage}\hfill
    \begin{minipage}{0.45\textwidth}
        \centering
        \includegraphics[width=0.9\textwidth]{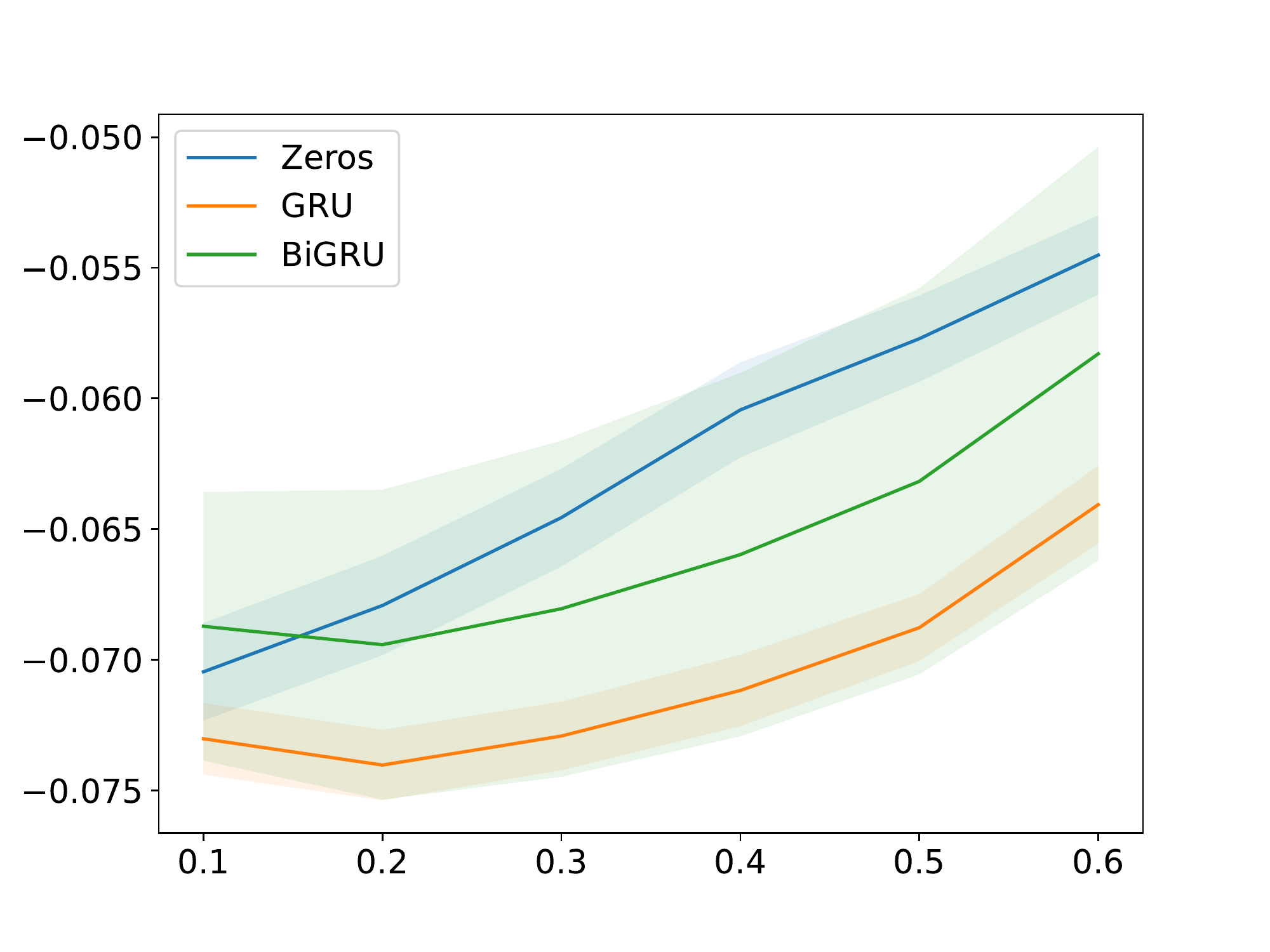} 
        \caption{Sufficiency results, masking between 10\% and 60\% of the data for each patient, and replacing the masked data with zeros: $\overline{x}_{t,i} = 0$. Lower is better with this metric.}
        \label{fig:suff_zeros_abl}
    \end{minipage}
\end{figure}

We also provide here more results of the ablation study, comparing using Zeros, a GRU or a BiGRU as a perturbation model. Results in terms of accuracy, cross-entropy, comprehensiveness and sufficiency can be found on Figures \ref{fig:acc_avg_abl}, \ref{fig:acc_zeros_abl}, \ref{fig:comp_avg_abl}, \ref{fig:comp_zeros_abl}, \ref{fig:suff_avg_abl} and \ref{fig:suff_zeros_abl}.

\begin{figure}[h]
    \centering
    \begin{minipage}{0.45\textwidth}
        \centering
        \includegraphics[width=0.9\textwidth]{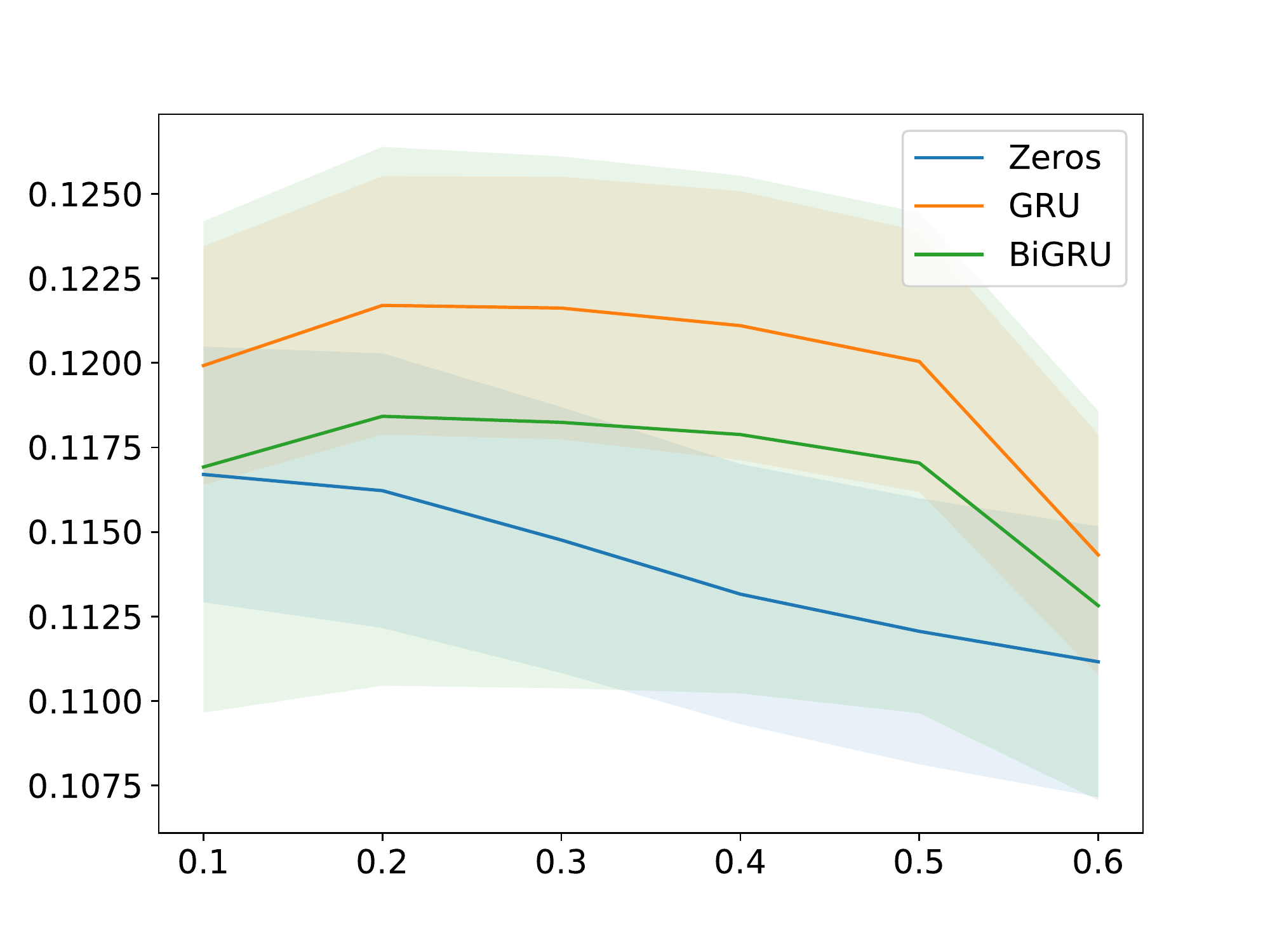} 
        \caption{Cross-entropy results, masking between 10\% and 60\% of the data for each patient, and replacing the masked data with the overall average over time for each feature: $\overline{x}_{t,i} = \frac{1}{T}\sum_{t} x_{t, i}$. Higher is better with this metric.}
        \label{fig:ce_avg_abl}
    \end{minipage}\hfill
    \begin{minipage}{0.45\textwidth}
        \centering
        \includegraphics[width=0.9\textwidth]{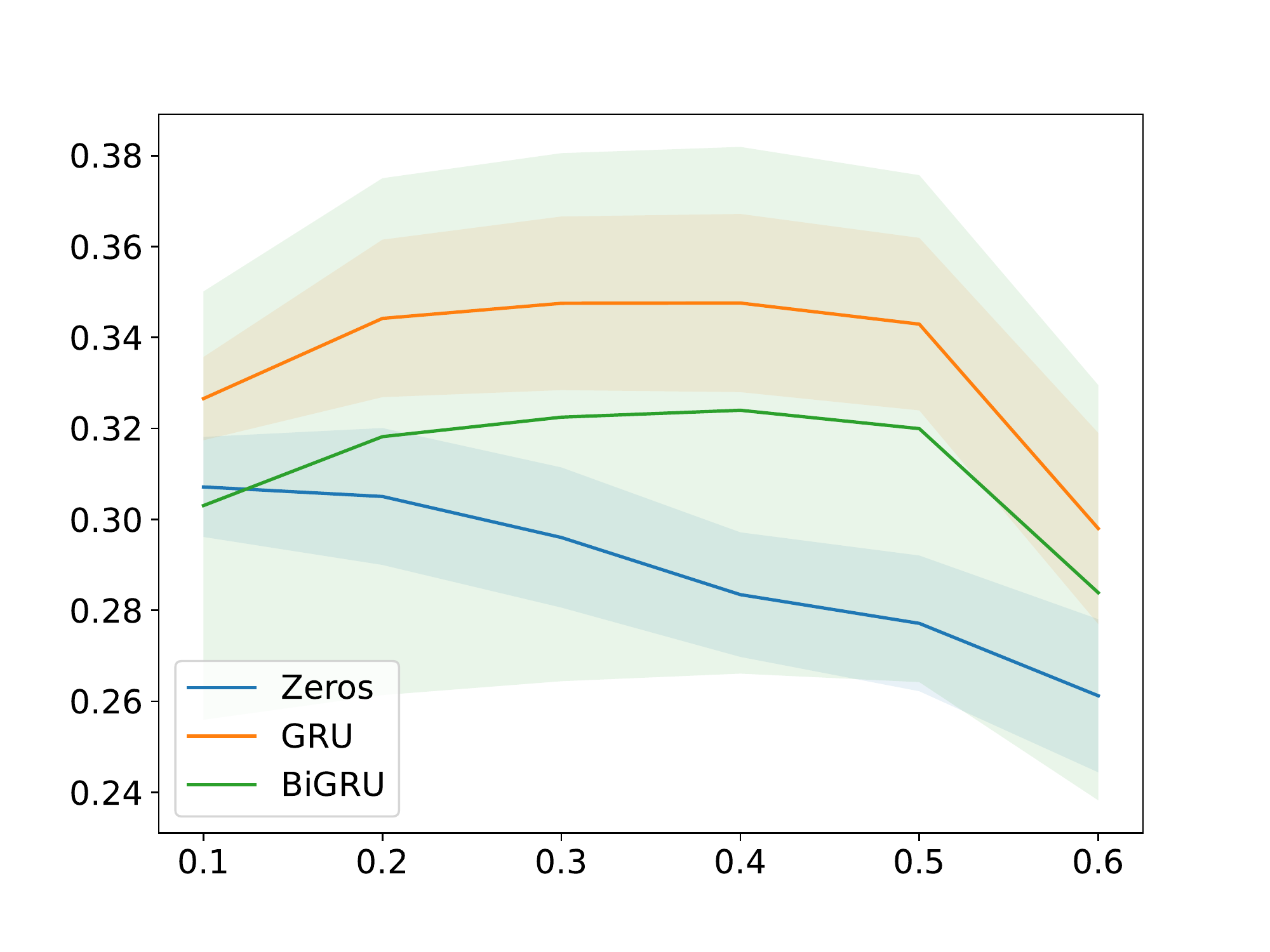} 
        \caption{Cross-entropy results, masking between 10\% and 60\% of the data for each patient, and replacing the masked data with zeros: $\overline{x}_{t,i} = 0$. Higher is better with this metric.}
        \label{fig:ce_zeros_abl}
    \end{minipage}
\end{figure}

\end{document}